\documentclass[journal]{IEEEtran}
\usepackage{float}
\usepackage{amsmath}
\newtheorem{myDef}{Definition}
\usepackage{subcaption}
\usepackage{amssymb}
\usepackage{CJK}

\usepackage{enumitem}
\usepackage{amsfonts}
\usepackage{multirow}
\usepackage{cite}
\usepackage{epstopdf}
\usepackage{graphicx}
\usepackage{float}
\usepackage{booktabs}
\usepackage{threeparttable}
\usepackage{stfloats}
\usepackage{algorithm}
\usepackage{algpseudocode}
\usepackage[justification=centering]{caption}
\usepackage[anchorcolor=blue,urlcolor=blue,linkcolor=blue,citecolor=black]{hyperref}
\usepackage{makecell}
\usepackage[numbers]{natbib}
\def\BibTeX{{\rm B\kern-.05em{\sc i\kern-.025em b}\kern-.08em
    T\kern-.1667em\lower.7ex\hbox{E}\kern-.125emX}}
\begin{document}
\title{Short-Term Multi-Horizon Line Loss Rate Forecasting of a Distribution Network Using Attention-GCN-LSTM}
\author{Jie~Liu,~\
       Yijia~Cao, ~\IEEEmembership{Senior Member,~IEEE,}
       Yong~Li, ~\IEEEmembership{Senior Member,~IEEE,}
       Yixiu~Guo,~\
       and~Wei~Deng,~\
       \thanks{J. Liu ,Y. Cao ,Y. Li  and  Y. Guo are with the College of Electrical and Information Engineering, Hunan University, Changsha 410082, China. (Corresponding author: Y. Cao, Y. Li. E-mail: yjcao@hnu.edu.cn ,yongli@hnu.edu.cn )  .}
\thanks{W. Deng is with the State Grid Hunan Electric Power Company Limited Research Institute, Changsha, 410007, China. }}

\maketitle

\begin{abstract}
Accurately predicting line loss rates is vital for effective line loss management in distribution networks, especially over short-term multi-horizons ranging from one hour to one week. In this study, we propose Attention-GCN-LSTM, a novel method that combines Graph Convolutional Networks (GCN), Long Short-Term Memory (LSTM), and a three-level attention mechanism to address this challenge. By capturing spatial and temporal dependencies, our model enables accurate forecasting of line loss rates across multiple horizons. Through comprehensive evaluation using real-world data from 10KV feeders, our Attention-GCN-LSTM model consistently outperforms existing algorithms, exhibiting superior performance in terms of prediction accuracy and multi-horizon forecasting. This model holds significant promise for enhancing line loss management in distribution networks.
\end{abstract}

\begin{IEEEkeywords}
Distribution Network Loss Forecasting, Attention Mechanism, Graph Convolutional Network(GCN), Long Short-Term Memory(LSTM).
\end{IEEEkeywords}

\section{Introduction}
\label{sec:introduction}
\IEEEPARstart{T}{HE} line loss rate is a crucial technical and economic indicator used to measure the management level of power supply enterprises. The distribution network's line loss is a vital element of the overall power grid, contributing significantly to its overall losses \cite{khodr2002standard}. Currently, managing line loss rates in distribution networks typically involves calculating and analyzing the line loss rate using basic statistics. However, this approach does not provide a comprehensive understanding of the critical factors influencing distribution network losses or the anticipated trends in line loss rates in the future. As a result, specific indicators to guide the reduction of losses in the distribution network are lacking \cite{jin2021research, ni2019review}. Therefore, there is a need for a more comprehensive line loss management analysis to identify the key factors that impact distribution network losses and provide more targeted guidance for reducing line losses in the future.

Advancements in smart grid technology and the widespread adoption of smart meters, coupled with the availability of real-time power databases and large-scale data platforms, have opened up new avenues for utilizing techniques such as data mining, machine learning, and deep learning to identify crucial factors that impact the loss rate of distribution networks \cite{ekanayake2012smart}. Implementing short-term multi-horizon line loss rate prediction enables the determination of the future line loss fluctuation range, which can aid power supply enterprises in making optimal decisions for loss reduction. This transformation shifts power supply enterprises from a passive operational model to an active forecasting model and establishes a new decision-making model driven by big data, namely "correlation + prediction + regulation" \cite{zhang2018big, liu2023analysis}. By leveraging this approach, power supply enterprises can effectively plan and implement loss reduction strategies, which can lead to improved operational efficiency and reduced costs.
\begin{figure*}[hb]
\centering
\includegraphics[scale=0.5]{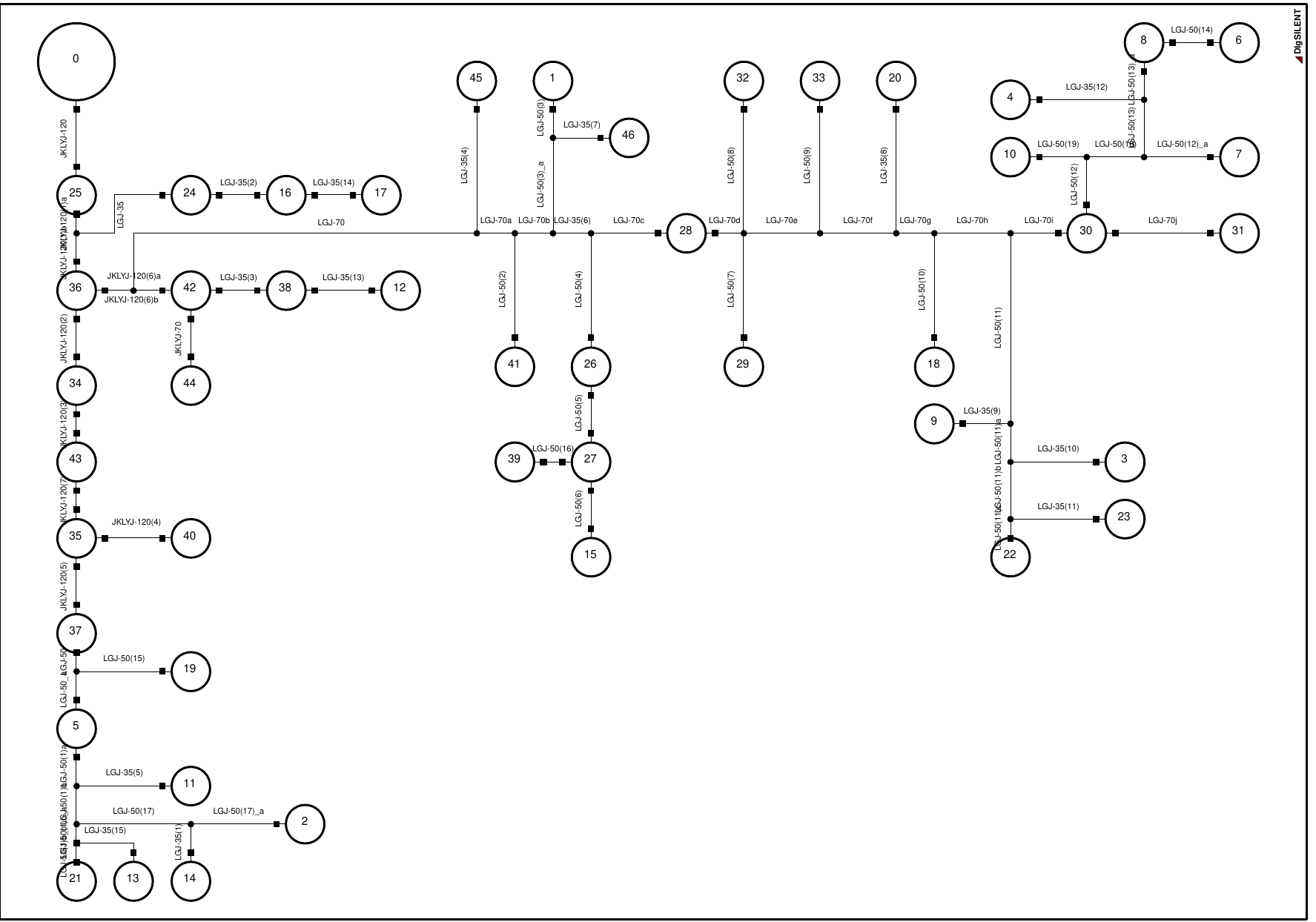}

\caption{Feeder Topology in a Distribution Network. The feeder topology illustrated in the figure showcases a distribution network with 44 substations. Each label circle represents a substation, and within each substation, there is a 10/0.4 kV transformer along with an equivalent load on the low-voltage side. The type of feeder branch is indicated by by the characters and numbers displayed on each line.}
\label{fig:caizhenline}
\end{figure*}

The existing methods for predicting line loss rate in distribution power networks can be categorized into two main groups: physics-based models and data-driven models. Physics-based models are based on the fundamental physical laws that govern the flow of electricity in power systems \cite{sun1980calculation, fan2016research, hu2021real}. Physics-based models utilize mathematical equations to simulate the behavior of power systems and forecast line losses. However, they require detailed knowledge of the network topology, system parameters, and operating conditions. Moreover, due to their complex calculations and simulations, physics-based models can be computationally intensive. On the other hand, data-driven models rely on statistical and machine learning techniques to discover patterns and relationships in historical data, which are then utilized for making predictions \cite{min2015application, wang2017line, yao2019research, zhang2019research, yitao201910, tulensalo2020lstm, dalal2021day, liu2023analysis}. Data-driven models have the advantage of being trained on large datasets and can adapt well to changing conditions. However, they may struggle to capture all the intricate interactions between different components of the power system, potentially limiting their prediction accuracy.

Forecasting the line loss rate in distribution power networks is a complex task, primarily due to the intricate spatial and temporal dependencies involved.that has always been challenging due to the intricate spatial and temporal dependencies that are inherently involved. The distribution power network can be represented as a graph, where the topological structure plays a crucial role \cite{kim2019graph}. Fig.~\ref{fig:caizhenline} illustrates this structure, where a feeder consists of 44 transformer districts. Each node in the graph represents a transform district, which includes a distribution transformer and an equivalent load on the low-voltage side. The electrical parameters of these nodes, representing their states, continuously evolve due to the influence of neighboring nodes and other factors until reaching an equilibrium state. The spatial relationship between neighboring transformer districts plays a significant role in determining the overall line loss of this distribution network. Furthermore, the future overall line loss of this distribution network is influenced not only by the historical electrical states but also by dynamic external factors such as feeder type, transformer type, and weather conditions \cite{georgilakis2015review}.

In the field of forecasting line loss rate in distribution networks, several challenges need to be addressed. Firstly, previous studies have relied on indirect prediction methods using artificial neural networks and genetic algorithms, which can lead to error accumulation when making predictions based on load or generation forecasts \cite{min2015application, dalal2021day}. Secondly, most existing models (\cite{min2015application, wang2017line, yao2019research, zhang2019research, yitao201910, tulensalo2020lstm, dalal2021day, liu2023analysis}) only provide single-scale predictions, making it difficult to track the progress of line loss rate, identify anomalous states, and quickly adjust operational modes. This limitation hinders the ability to forecast line loss rate accurately. Thirdly, the temporal correlation of measurement data from individual transformer districts has been considered in previous studies, but the spatial correlation between transformer districts and the integration of data from all transformer districts to predict line losses have been overlooked. This neglect of spatial correlation limits the models' ability to accurately forecast future line loss rate.

Given these challenges, it is crucial to develop an approach that explicitly addresses the spatial and temporal dependencies in the distribution network and integrates the data from all transformers to forecast line losses accurately. The proposed study aims to overcome these challenges and improve the accuracy and effectiveness of line loss rate predictions in distribution networks.

In their work, \cite{zhao2019t} proposed a model that combines graph convolutional networks (GCN) and long short-term memory (LSTM) networks to capture spatial and temporal interrelationships. To effectively model complex traffic patterns, their model incorporate a two-level attention mechanism, which assigns weights to different nodes and time steps. Building upon their approach, we have adapted this methodology to predict line loss rate in a distribution network. Our primary objective is to comprehensively capture the spatial and temporal relationships present in the recorded data for each distribution transformer, thereby significantly improving the accuracy of our predictions. Despite the computational expenses and susceptibility to overfitting associated with LSTM models, their effectiveness in capturing long-term dependencies and temporal patterns in various sequence-based prediction tasks has been well-established. Considering that the dataset of the power distribution network exhibits long-term dependencies and complex temporal patterns, LSTM models are considered more suitable for capturing these dynamics compared to alternative models. To mitigate the risk of overfitting and enhance the model's generalization capabilities, we have employed regularization techniques, such as dropout and L2 regularization.

In our research, we have taken into consideration the impact of electrical parameters, network characteristics and weather conditions on line loss prediction. By including these variables, we aim to capture the direct and indirect effects of facors on power system operations and line losses. The additional electrical features provide important information about the electrical characteristics of the distribution network, such as voltage levels, power factor, and load profiles. These features enable the model to understand the electrical conditions of the network and their influence on line losses. Furthermore, the inclusion of weather metrics such as temperature, humidity, wind speed, and precipitation allows our model to account for the external environmental factors that can impact power distribution. High temperatures, for example, can lead to increased electrical load and higher line losses, while extreme weather events such as storms or strong winds can cause disruptions and damage to the distribution infrastructure, resulting in increased line losses \cite{georgilakis2015review}. By integrating these electrical features and weather metrics into our model, we create a more comprehensive and accurate framework for line loss prediction. This approach enables our model to capture the complex relationship between weather conditions, electrical characteristics, and line losses, resulting in improved forecasting performance.

To summarize, our contributions can be outlined as follows:

\begin{enumerate}
\item Integration of Feeder Topology and SCODA Data: In a novel approach, we  are the first to combine feeder topology and SCODA (Supervisory Control and Data Acquisition) data to predict the line loss rate in the distribution network, enabling more accurate predictions by incorporating both physical structure and real-time operational data.
\item Comprehensive Input Features: To capture the diverse factors influencing line loss, we incorporate multiple electrical and non-electrical indicators, including meter measurements, line characteristics, transformer capacity. Additionally, we integrate weather data, considering its significant impact on power generation, transmission, and load.
\item Enhanced Attention Mechanism: We utilize a three-level attention mechanism to enhance line loss rate prediction accuracy by evaluating the contribution of each transformer district, assessing the importance of latent features, and learning the relevance of information at each time step.
\item Multi-Horizon Line Loss Forecasting: Our developed model accurately forecasts line loss rates across different time horizons, from short-term (one-hour) to long-term (one-week) predictions, demonstrating its effectiveness in revealing patterns and dependencies for line loss management in the distribution network.

\end{enumerate}

We evaluated our approach using actual sample data from 44 distribution transformer districts in a 10KV distribution network in LingLing, Hunan province. Our proposed method demonstrated superior performance compared to baseline methods. The Attention-GCN-LSTM model consistently outperformed all baselines across various evaluation metrics and prediction horizons. Notably, in a one-week forecast, our model maintained strong performance with a 0.7687 R2 score, which was 14.54\% higher than the best baseline result. Additionally, the model reduced the RMSE and MAE by 23.64\% and 21.77\%, respectively.These results provide strong evidence of the effectiveness of the Attention-GCN-LSTM model for line loss rate forecasting.

\section{RELATED WORK}
\subsection{Data-driven Methods in Line Loss Prediction}

Previous research on data-driven line loss prediction in distribution networks can be classified into two primary categories: traditional time series forecasting models and machine learning models. Traditional models, such as autoregressive moving average (ARMA) and autoregressive integrated moving average (ARIMA), have been commonly used for line loss prediction. These models rely on historical data patterns to forecast future line loss rates. Machine learning models, including support vector machines (SVM), decision tree models (such as gradient boosting decision trees or random forest), and neural network models (such as backpropagation and recurrent neural networks), have also been applied in line loss prediction.

In one study \cite{min2015application}, a particle swarm optimization (PSO)-based support vector regression (SVR) algorithm was employed to forecast low voltage line loss rates for the next three years. The study focused on a limited number of factors influencing line loss and made long-term predictions. Another approach, presented in \cite{wang2017line}, utilized a hierarchical clustering algorithm to group distribution transformers and developed Random Forest (RF) estimation models for different transformers. However, the prediction range of random forest models is limited by the maximum and minimum values in the training data, which can lead to inaccurate predictions when the data distribution changes.

To address these limitations,  \cite{yitao201910} introduced an improved backpropagation (BP) neural network method for line loss prediction. Despite efforts to enhance the BP network, challenges such as slow convergence and vulnerability to local minima still exist. The representativeness of the training samples plays a crucial role in the model's approximation and generalization abilities, making it challenging to select representative samples accurately \cite{liu2023analysis}.

In the work of Zhang et al. (2019) \cite{zhang2019research}, a backpropagation (BP) neural network model was constructed using the LM numerical optimization algorithm and K-means clustering technique to forecast line loss rates for individual transformer districts. Another study by \cite{tulensalo2020lstm} employed a bidirectional LSTM model to predict power grid losses. This architecture processes the input sequence in both forward and backward directions, capturing a more comprehensive representation of the data. Additionally, \cite{dalal2021day} used CatBoost, an open-source gradient boosting library, to forecast grid loss for each hour of the day-ahead. The features used in these studies included load predictions, weather forecasts, and calendar features. \cite{yao2019research} proposed a line loss rate prediction approach based on gradient boosting decision trees (GBDT). However, this method only considered a limited number of electrical and non-electrical features. However, some of these approaches had a limited selection of electrical parameters, and using predicted load data as features may introduce error accumulation problems.

Furthermore, \cite{liu2023analysis} employed machine learning techniques to examine the associations between line loss and nine electrical and non-electrical variables. They identified parameters with strong correlations and used them as inputs for a forecasting model based on the Long Short-Term Memory (LSTM) architecture. This approach enabled them to predict line loss for the next day in eight different time intervals, each spanning three hours.

\begin{table*}[ht]
\centering
\caption{Survey of the Benchmark Forecasting Methods for Line Loss Prediction.}
\resizebox{\textwidth}{28mm}{
\begin{tabular}{|c|l|l|}
\hline
\textbf{Methods} & \multicolumn{1}{c|}{\textbf{Advantages}}                                                                                                                                                                                                                                                                      & \multicolumn{1}{c|}{\textbf{Disadvantages}}                                                                                                                                                                                               \\ \hline
\textbf{ARIMA}   & \begin{tabular}[c]{@{}l@{}}Can capture trends and seasonality in time series data \cite{hyndman2018forecasting}.\\ Provides a straightforward way to incorporates historical data into predictions \cite{shumway2017arima}.\\ Has a simple and interpretable model structure \cite{brockwell2016arma}.\end{tabular}                                                                           & {\begin{tabular}[c]{@{}l@{}}Assumes that the data is stationary, which may not be the case in line loss rate prediction. \cite{hyndman2018forecasting}.\\ May underperform with complex dependencies and nonlinear relationships in line loss rate patterns.\cite{wei2006time}.\end{tabular}} \\ \hline
\textbf{BP}      & \begin{tabular}[c]{@{}l@{}} Can capture complex relationships between input and output variables \cite{goodfellow2016deep}.\\ Can handle both continuous and categorical input features \cite{geron2017hands}.\end{tabular}                                                                                                                                    & \begin{tabular}[c]{@{}l@{}}Prone to overfitting and slow convergence \cite{goodfellow2016deep}.\\Can be sensitive to the choice of hyperparameters \cite{goodfellow2016deep}.\\ May have difficulty capturing long-term dependencies in time series data, which can impact their accuracy in forecasting line loss rates. \cite{hochreiter1997long}.\end{tabular}                                                                    \\ \hline
\textbf{SVR}     & \begin{tabular}[c]{@{}l@{}}Relatively robust to noise and outliers in the input data \cite{muller2018introduction}.\\Good at handling high-dimensional data \cite{scholkopf2002learning}.\\Can capture nonlinear relationships between input and output variables \cite{scholkopf2002learning}.\\ Can handle both continuous and categorical input features \cite{smola2004tutorial}. \end{tabular}                                                                 & \begin{tabular}[c]{@{}l@{}}Requires careful tuning of its parameters which can be time-consuming and require domain expertise \cite{smola2004tutorial} .\\Can be sensitive to the choice of kernel function and hyperparameters \cite{scholkopf2002learning}.\\ May have difficulty capturing long-term dependencies in line loss rate patterns. \cite{chen2016xgboost}.\end{tabular}                                                \\ \hline
\textbf{RF}      & \begin{tabular}[c]{@{}l@{}}Can handle both continuous and categorical input features \cite{breiman2001random}.\\ Can capture complex relationships between input and output variables \cite{breiman2001random}.\\ Can handle missing data and outliers \cite{breiman2001random}.\\Can provide feature importance rankings \cite{breiman2001random}.\end{tabular}                                                    & \begin{tabular}[c]{@{}l@{}}Can suffer from overfitting if the model is too complex or the number of trees in the forest is too large \cite{breiman2001random}.\\Does not perform well when the time series data exhibits strong autocorrelation, which is common in line loss rate patterns. \cite{meinshausen2006quantile}.\\ May have difficulty capturing long-term dependencies in line loss rate patterns. \cite{cutler2007random}.\end{tabular}                                                                  \\ \hline
\textbf{GBDT}    & \begin{tabular}[c]{@{}l@{}}Can capture complex relationships between input and output variables \cite{chen2016xgboost}.\\ Can handle both continuous and categorical input features \cite{chen2016xgboost}.\\ Can provide feature importance rankings \cite{friedman2001greedy}.\\ Can handle missing data and outliers \cite{chen2016xgboost}.\end{tabular}                                                 & \begin{tabular}[c]{@{}l@{}}Susceptible to overfitting, especially when dealing with high-dimensional features or noisy data in line loss prediction \cite{chen2016xgboost}.  \\ Difficulty capturing long-term dependencies in line loss prediction \cite{lai2018modeling}. \\ Limited interpretability and difficult to identify the most important features for line loss prediction \cite{lundberg2017unified}.\end{tabular}                                                                  \\ \hline
\textbf{LSTM}    & \begin{tabular}[c]{@{}l@{}}Can capture long-term dependencies in time series data \cite{hochreiter1997long}.\\ Can handle nonlinearity and non-stationarity in time series data \cite{lipton2015learning}.\\ Can be trained end-to-end and can learn representations of the input data \cite{hochreiter1997long}.\end{tabular}            & \begin{tabular}[c]{@{}l@{}}Can be computationally expensive and time-consuming to train \cite{pascanu2013difficulty}.\\ Susceptible to overfitting and difficulty handling noise and outliers, which are common characteristics of line loss data. \cite{lipton2015learning}.\\ Still struggle with retaining information over very long time horizons in line loss data \cite{gers2000learning}.\end{tabular} \\ \hline
\textbf{Bi-LSTM} & \begin{tabular}[c]{@{}l@{}}Can capture both past and future information from the input sequence \cite{schuster1997bidirectional}.\\ Better handling of long-term dependencies, leading to more accurates forecasts. \cite{graves2013generating}.\end{tabular} & \begin{tabular}[c]{@{}l@{}}Increased computational complexity and less practical for in practical large-scale applications of line loss rate prediction\cite{greff2016lstm}.\\ May not be suitable for all types of line loss data, for example, if data exhibits cyclic patterns (e.g., daily or weekly patterns)\cite{chung2014empirical}. \\ Have a larger number of parameters than LSTMs and increase the risk of overfitting \cite{li2018stock}. \end{tabular} \\ \hline
\textbf{TCN-LSTM} & \begin{tabular}[c]{@{}l@{}} Allows for capturing both short-term dependencies and long-term patterns in line loss rate data. \cite{bi2021hybrid}.\\ The integration of TCN and LSTM enables the model to handle the complex and dynamic nature of line loss rate patterns. \cite{bi2021hybrid}.\end{tabular} & \begin{tabular}[c]{@{}l@{}}The computational complexity of the hybrid prediction method may be higher compared to individual TCN or LSTM models\cite{cheng2021high}.\\ The method may require a large amount of training data and careful parameter tuning to achieve optimal performance in line loss rate prediction.\cite{cheng2021high}.  \end{tabular} \\ \hline
\textbf{DBN} & \begin{tabular}[c]{@{}l@{}}Can capture intricate and nonlinear relationships present in time series data \cite{wang2019adaptive}.\\ Can automatically learn and extract relevant features from the input time series\cite{wang2019adaptive}. \\ Adaptability to varying data distributions\cite{wang2019adaptive}. \end{tabular} & \begin{tabular}[c]{@{}l@{}}Can be computationally expensive and time-consuming for line loss rate prediction\cite{hinton2009deep}.\\ Deep belief networks may face challenges in capturing long-term dependencies in line loss data\cite{hinton2009deep}. \\May require expertise and extensive experimentation to tune hyperparameters \cite{hinton2009deep}.\end{tabular} \\ \hline
\textbf{PE-CNN} & \begin{tabular}[c]{@{}l@{}}Position encoding enhances CNNs' ability to capture temporal dynamics in line loss rate patterns \cite{jin2022position}.\\Position encoding enhances CNNs' feature extraction for identifying relevant patterns and relationships in line loss rate data \cite{jin2022position}.\end{tabular} & \begin{tabular}[c]{@{}l@{}}Capturing long-term dependencies or temporal patterns beyond the receptive field is challenging for CNNs, even with position encodings.\cite{goodfellow2016deep}.\\ Sensitive to input data quality: noisy, incomplete, or poorly formatted input data can negatively impact the performance\cite{goodfellow2016deep}.  \end{tabular} \\ \hline
\end{tabular}}
\label{Tab:Survey}
\end{table*}

\subsection{State-of-the-Art Methods for Time Series Prediction}

As line loss rate prediction involves time series data, it is essential to explore and apply the latest advancements in time series prediction methods to enhance the accuracy and effectiveness of forecasting in distribution networks. Recent research has introduced state-of-the-art techniques that offer potential improvements in line loss rate prediction.

One such method (TCN-LSTM) proposed in\cite{bi2021hybrid} is the hybrid prediction method that combines temporal convolutional networks (TCN) and long short-term memory (LSTM) networks to forecast realistic network traffic. This method addresses the challenge of capturing both short-term dependencies and long-term patterns in network traffic data. By integrating TCN and LSTM, the model can capture temporal dependencies at different scales, resulting in improved prediction accuracy and robustness.

In \cite{wang2019adaptive}, an adaptive deep belief network (DBN) with sparse restricted Boltzmann machines (RBMs) is introduced. This method combines the strengths of DBNs and sparse RBMs to enhance the learning capabilities of the network and address the limitations of traditional deep learning models. The adaptive DBN exhibits superior performance and adaptability in various tasks, presenting a powerful and flexible framework for modeling complex data distributions.

Another notable approach (PE-CNN) proposed in \cite{jin2022position} is the use of Convolutional Neural Networks (CNNs) with position encoding for predicting the remaining useful life (RUL) of machine systems. This approach incorporates position encoding techniques to capture the temporal dynamics in time series data, enabling the CNN to extract relevant features for accurate RUL prediction. Experimental results demonstrate the effectiveness of this approach, showcasing its superiority over baseline models.

We can consider both previous research on data-driven line loss prediction in distribution networks and state-of-the-art methods for time series prediction as benchmark models. A comprehensive comparison of these benchmark forecasting methods is presented in Table ~\ref{Tab:Survey}.

\section{METHODOLOGY}
In this section, we present our proposed Attention-GCN-LSTM model for line loss prediction in power distribution networks. This model combines the strengths of LSTM for temporal dependency learning and graph convolutional networks for capturing spatial dependence based on the distribution network's topology \cite{bai2021a3t}. Moreover, we introduce a three-level attention mechanism that adjusts the importance of various distribution transformers, latent features, and time points, integrating global spatial-temporal information to enhance prediction accuracy. For simplicity, we will not consider the batch dimension throughout the presentation.
\begin{figure*}[htb]
\centering
\includegraphics[scale=0.45]{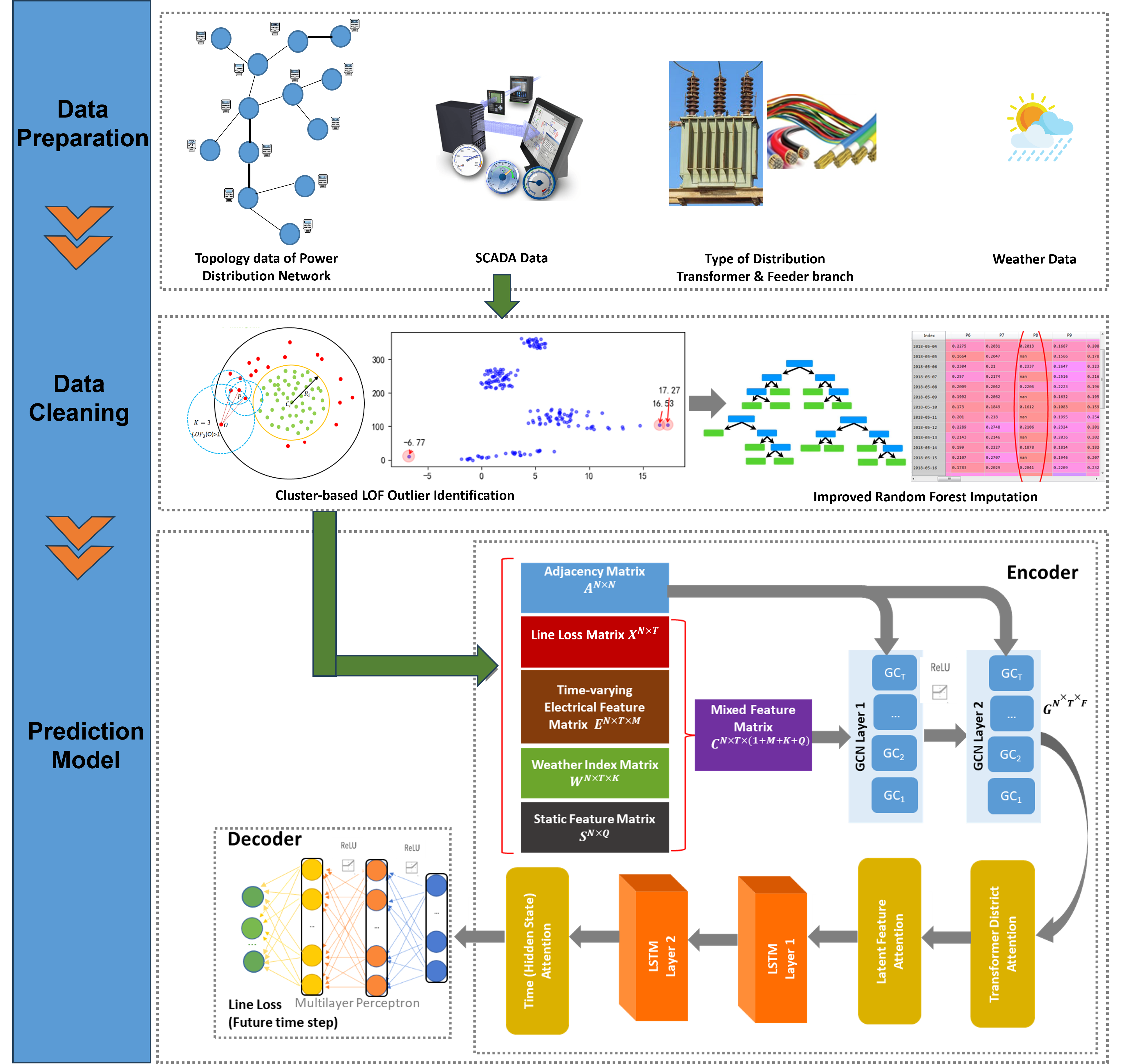}

\caption{The Overall Framework for Line Loss Prediction in Power Distribution Networks. The framework consists of three main components: data preparation, data cleaning, and prediction model. The data preparation component involves gathering network topology, SCADA data, details about the main network feeder, distribution transformers, branch lines, and weather data. The data cleaning component addresses abnormal or missing data using the cluster-based LOF algorithm and the Improved Random Forest algorithm. The prediction model utilizes the cleaned historical SCADA data, network topology, network characteristics, and weather data to forecast the future overall line loss. It combines LSTM for temporal dependency learning and graph convolutional networks for spatial dependence based on the network topology. The model incorporates a three-level attention mechanism to adjust the importance of transformers, latent features, and time points, enhancing prediction accuracy.}
\label{fig:framwork}
\end{figure*}

\subsection{Problem Definition}
As depicted in Fig. \ref{fig:framwork}, the comprehensive system framework consists of three main components. The first component is data preparation, which involves gathering the network topology of the distribution network feeder, SCADA data from each transformer district, details about the main network feeder, distribution transformers, branch lines from each transformer district, and weather data. The second component is data cleaning, which focuses on addressing abnormal or missing data in SCADA records caused by various factors such as equipment failures, noise interference, transmission errors, and abnormal power consumption. Lastly, the prediction model utilizes the cleaned historical SCADA data, network topology, network characteristics, and weather data to forecast the future overall line loss of the feeder. The overall loss of the feeder plays a critical role in power system management and optimization \cite{liu2023analysis}. The main objective of this study is to develop a reliable prediction model that can accurately forecast the future overall loss of the feeder using historical data. To facilitate understanding, the symbols, feature matrix representations, and line loss prediction functions are defined as follows:

\begin{myDef}
Feeder network G. For a given feeder in a distribution network, its topological structure, static properties, and temporal properties are described as graph \begin{math} G = (V,E) \end{math}, with a set of vertices \begin{math} V =\{v_1,v_2,v_3,...,v_N\}\end{math} where \begin{math}N\end{math} is the number of transformer districts, and a set of edges \begin{math} E \end{math} indicating the connections between them. Each vertex in the graph represents a transformer district. The adjacency matrix A is a square matrix of size N x N, where each element represents the existence of a directed edge between vertex i and vertex j. Specifically, if there is an edge from vertex i to vertex j, then A(i,j) = 1; otherwise, A(i,j) = 0.
\end{myDef}
\begin{myDef}
   Line loss matrix $X$, of size \begin{math}{N\times T}\end{math}, is constructed, where each row represents a transformer district and each column represents a time series. Since it is impractical to calculate the branch line loss of each transformer district that contains thousands of user meters, we use the historical overall line loss of the entire feeder as a feature for each transformer district. To create this feature, we expand the historical overall line loss matrix, denoted as \begin{math}L\in\end{math}, which is of size  \begin{math} 1\times T \end{math}, \begin{math}N\end{math} times to generates a matrix \begin{math}X\in\end{math} of size \begin{math}N\times T\end{math} in the real numbers, denoted as \begin{math}X\in\end{math}\begin{math}\mathbb{R}^{N\times T} \end{math}. This feature matrix $X$ also serves as the target variable that the model aims to predict.
\end{myDef}
\begin{myDef}
  The time-varying electrical feature matrix $E$, denoted as \begin{math}E\in\end{math} \begin{math} \mathbb{R}^{N\times (T\times M)} \end{math}, is created to store all the time-series data related to voltage, current, power, and other electrical parameters. Here, \begin{math}N\end{math} represents the number of transformer districts, \begin{math}T\end{math} represents the length of the historical time series, and \begin{math}M\end{math} is the total number of time-varying electrical features. The matrix \begin{math}E\end{math} is represented as \begin{math} E ={E_{i1},E_{i2},E_{i3},...,E_{iM}}\end{math}, where \begin{math}E_{ij}\end{math} represents the \begin{math}j\end{math}th time-varying electrical feature of the \begin{math}i\end{math}th transformer district. Each \begin{math}E_{ij}\end{math} is defined as \begin{math}E_{ij} ={j_i^{t-T},j_i^{t-T+1},...,j_i^{t-1}}\end{math}, where \begin{math}j_i^t\end{math} denotes the \begin{math}j\end{math}th time-varying electrical feature of the \begin{math}i\end{math}th transformer district at time \begin{math}t\end{math}.
\end{myDef}
\begin{myDef}
   Static feature matrix \begin{math}S^{N\times Q} \end{math}. The type and capacity of the transformer, as well as the length and type of the line, may differ among distribution transformers. However, these attributes remain static and do not change over time. Therefore, we represent them as a static feature matrix \begin{math}S\in\end{math} \begin{math} \mathbb{R}^{N\times Q} \end{math}, where \begin{math}Q\end{math} represents the number of static features, corresponding to the total number of categories obtained through one-hot encoding of the static attributes. The values in the static feature matrix \begin{math}S\end{math} are numerical, with each column representing a specific category of the static attribute.
\end{myDef}
\begin{myDef}
   Weather index matrix \begin{math}W^{N\times T \times K} \end{math} . Weather-related factors such as temperature, humidity, wind direction, wind force, solar intensity, etc., vary over time. To capture these variations, we represent them as a time-varying weather index matrix denoted by \begin{math}W\in\end{math} \begin{math} \mathbb{R}^{N\times T \times K} \end{math}, where \begin{math}T\end{math} represents the length of the historical time series, and \begin{math}K\end{math} represents the number of weather indexes. Each element of the matrix \begin{math}W\end{math} corresponds to a specific weather index at a particular time for a given transformer district. \end{myDef}

\begin{myDef}
 Mixed feature matrix \begin{math}C \end{math}. The input features for each node (distribution transformer) in the feeder's topology \begin{math}G\end{math} comprise historical line loss matrix \begin{math}X\end{math}, time-varying ancillary feature matrix \begin{math}E\end{math}, statical feature matrix \begin{math}S\end{math} (expanded to T time steps), and weather index matrix \begin{math}W\end{math}. These features are merged to form a new matrix \begin{math}C\in\end{math} \begin{math} \mathbb{R}^{N\times T\times (1+M +Q+ K)} \end{math}.
 The mixed feature matrix \begin{math}C \end{math} combines the input features for each node in the feeder's topology, including the historical line loss matrix $X$, time-varying ancillary feature matrix $E$, statical feature matrix $S$ (expanded to $T$ time steps), and weather index matrix $W$. The matrix $C$ is of size \begin{math} N\times T\times (1+M +Q+ K) \end{math} where $N$ represents the number of transformer districts, $T$ represents the length of the historical time series, $M$ represents the total number of time-varying electrical features, $Q$ represents the number of static features, and $K$ represents the number of weather indexes. Each element in matrix $C$ corresponds to a specific feature value for a given transformer district at a particular time. This merging step allows for a comprehensive representation of the input features, capturing both temporal and static information, and enabling the prediction model to learn the complex relationships between these features and the target variable.
\end{myDef}

The line loss prediction problem can be formulated as a structural sequence modeling task, aiming to learn a mapping function \begin{math}f\end{math} that takes the adjacency matrix \begin{math}A\end{math} of the distribution network \begin{math}G\end{math} and the mixed feature matrix \begin{math}C\end{math} as inputs to predict the future line loss rate \begin{math}\hat{y}\end{math}. The predicted line loss rates for \begin{math}P\end{math} time steps \begin{math}\hat{y}=[X_t, X_{t+1},..., X_{t+P-1}]\end{math} are obtained through the mapping function, as shown in Equation ~\ref{eq:equation1}:

\begin{equation}
\hat{y} = f(G, C)
\label{eq:equation1}
\end{equation}

To optimize the line loss prediction, we define a line loss function as follows:

\begin{equation}
\min_{f} \frac{1}{N} \sum_{i=1}^{N} (f(G, C_i) - y_i)^2
\end{equation}

where N is the total number of samples, \begin{math}f(G, C_i)\end{math} represents the predicted line loss rate for sample i based on the feeder topology G and the mixed feature matrix \begin{math}C_i\end{math}, and \begin{math}y_i\end{math} represents the true line loss rate for sample i. The objective is to minimize the mean squared difference between the predicted and true line loss rates.

\subsection{Data Preparation}

A feeder in the distribution network consists of multiple distribution transformers that form a network. The SCADA (Supervisory Control and Data Acquisition) system contains various parameters such as three-phase current, voltage, active and reactive power, and power consumption, which are collected from the Transformer Terminal Units (TTUs) installed in each transformer. These parameters are stored in a table where each row represents a day, and the columns correspond to 15-minute time intervals. The power consumption data is recorded once a day in a single column, while other parameters are recorded every 15 minutes, resulting in 96 columns per sample. These measurements are utilized to calculate important parameters such as load rates, power factor, and three-phase imbalance degree. The load rate is a measure of the efficiency of electrical energy usage, calculated by dividing the actual power consumed by the maximum power that could be consumed. The power factor is the ratio of the active power (representing the actual power used) to the apparent power (which is the product of current and voltage). A low power factor indicates inefficient use of power, potentially leading to higher energy costs. The three-phase imbalance degree measures the consistency in amplitude of the three-phase current or voltage in the power system, calculated by comparing the magnitudes of the three phases. In addition to these calculations, the PowerFactory software is employed to create a simulation model of the 10KV distribution network and determine the overall loss of the feeder.

We utilize the Meteostat Python library, which provides a convenient API for accessing open weather and climate data, to obtain hourly weather data. The historical observations and statistics available through Meteostat are collected from various reliable sources, including national weather services such as the National Oceanic and Atmospheric Administration (NOAA) and Germany's national meteorological service (DWD). These data sources ensure the quality and accuracy of the weather information used in our analysis.

The network topology and network characteristics, including information about the main network feeder, distribution transformers, and branch lines from each transformer district, are obtained from the State Grid Hunan Electric Power Company Limited. This reliable source provides accurate and comprehensive data that forms the basis for analyzing and modeling the distribution network.

All the aforementioned data will serve as inputs for the prediction model, as outlined in Table \ref{Tab:input}.
\begin{table}[]
\centering
\caption{Input Data of the Model. This table presents three types of information used as input for the model: temporal information (Historical Time-varying Electrical Features and Time-varying Weather Index), spatial information (Topology Structure), and network characteristics (Statistical Features). These data types collectively contribute to the accurate prediction of line loss rates in the distribution network.}
\begin{tabular}{|c|clll|}
\hline
\multirow{12}{*}{\begin{tabular}[c]{@{}c@{}}Historical\\ Time-varying\\  Electrical\\ Features\end{tabular}} & \multicolumn{1}{c|}{\multirow{3}{*}{Voltage}}       & \multicolumn{3}{c|}{Phase-A}              \\ \cline{3-5}
                                                                                                             & \multicolumn{1}{c|}{}                               & \multicolumn{3}{c|}{Phase-B}              \\ \cline{3-5}
                                                                                                             & \multicolumn{1}{c|}{}                               & \multicolumn{3}{c|}{Phase-C}              \\ \cline{2-5}
                                                                                                             & \multicolumn{1}{c|}{\multirow{3}{*}{Current}}       & \multicolumn{3}{c|}{Phase-A}              \\ \cline{3-5}
                                                                                                             & \multicolumn{1}{c|}{}                               & \multicolumn{3}{c|}{Phase-B}              \\ \cline{3-5}
                                                                                                             & \multicolumn{1}{c|}{}                               & \multicolumn{3}{c|}{Phase-C}              \\ \cline{2-5}
                                                                                                             & \multicolumn{4}{c|}{\begin{tabular}[c]{@{}c@{}}Three-phase total\\ active power\end{tabular}}   \\ \cline{2-5}
                                                                                                             & \multicolumn{4}{c|}{\begin{tabular}[c]{@{}c@{}}Three-phase total\\ reactive power\end{tabular}} \\ \cline{2-5}
                                                                                                             & \multicolumn{4}{c|}{\begin{tabular}[c]{@{}c@{}}Three-phase \\ imbalance\end{tabular}}           \\ \cline{2-5}
                                                                                                             & \multicolumn{4}{c|}{Power factor}                                                               \\ \cline{2-5}
                                                                                                             & \multicolumn{4}{c|}{Load rates}                                                                  \\ \cline{2-5}
                                                                                                             & \multicolumn{4}{c|}{Total loss rates of the feeder}                                                                  \\ \hline

\multirow{7}{*}{\begin{tabular}[c]{@{}c@{}}Time-varying\\ Weather Index\end{tabular}}                        & \multicolumn{4}{c|}{Temperature}                                                                \\ \cline{2-5}
                                                                                                             & \multicolumn{4}{c|}{Humidity}                                                                   \\ \cline{2-5}
                                                                                                             & \multicolumn{4}{c|}{Wind direction}                                                             \\ \cline{2-5}
                                                                                                             & \multicolumn{4}{c|}{Wind speed}                                                                 \\ \cline{2-5}
                                                                                                             & \multicolumn{4}{c|}{Sunhour}                                                                    \\ \cline{2-5}
                                                                                                             & \multicolumn{4}{c|}{Visibility}                                                                 \\ \cline{2-5}
                                                                                                             & \multicolumn{4}{c|}{DewPoint}                                                                   \\ \hline

\multicolumn{1}{|l|}{\multirow{1}{*}{Topology Structure}}                                                     & \multicolumn{4}{c|}{The adjacency matrix of transformer districts }                                                           \\ \hline
\multicolumn{1}{|l|}{\multirow{2}{*}{Statical Features}}                                                     & \multicolumn{4}{c|}{Type of transformer}                                                           \\ \cline{2-5}
\multicolumn{1}{|l|}{}                                                                                       & \multicolumn{4}{c|}{Type of feeder branch}                                                                \\ \hline
\end{tabular}
\label{Tab:input}
\end{table}

\subsection{Data Cleaning}

\begin{figure*}[hb]
\centering
\includegraphics[scale=0.39]{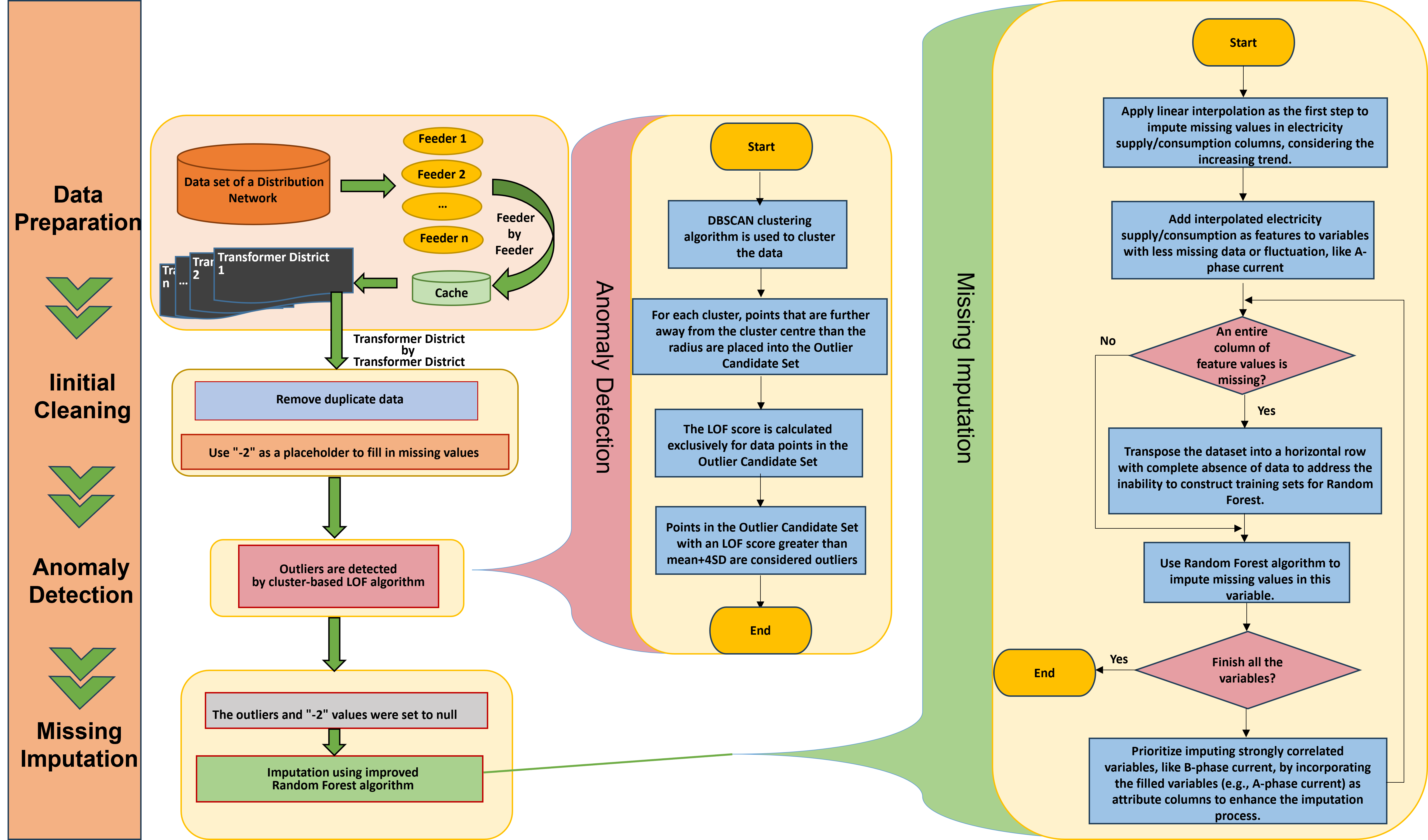}

\caption{Data Cleaning Framework for Distribution Network. The data cleaning framework for distribution networks comprises various techniques and processes aimed at enhancing data quality and reliability. This framework encompasses crucial steps, including Anomaly Detection and Missing Data Imputation. Anomaly Detection involves the utilization of the CLOF (Cluster-based Local Outlier Factor) algorithm, which enhances the LOF method by employing the K-means clustering algorithm for data clustering. On the other hand, Missing Data Imputation adopts an Improved Random Forest algorithm capable of effectively handling different forms of missing data, including blocked missing data and completely missing horizontal or vertical data.}
\label{fig:dataclean}
\end{figure*}

The data collected by SCADA can become abnormal or missing due to a range of factors such as measurement equipment failures, noise interference, errors during data transmission, and abnormal power consumption. Within our experimental dataset, the term ``bad data" encompasses both missing data points and outliers that have been identified using our outlier detection algorithm. It is important to note that these outliers and missing values are not artificially introduced, but rather reflect real-world occurrences that can significantly impact the analysis \cite{chen2020sensing}. In total, these problematic data points, which consist of both missing values and outliers, constitute approximately 9.59\% of the entire dataset. Therefore, data cleaning is necessary to identify and correct any anomalies or missing values. To address this, as shown in Fig.~\ref{fig:dataclean}, we employed a data cleaning process that involved identifying outliers using a cluster-based LOF algorithm and filling in missing data using an improved Random Forest algorithm \cite{liu2020big}. This approach helped to maintain the integrity of the data and avoid the potential negative impact on forecast accuracy that can occur when missing values are simply removed.

\textbf{\textit{Anomaly Detection}}
We utilize the cluster-based LOF (Local Outlier Factor) algorithm \cite{liu2020big} for outlier detection. During the outlier detection process, the LOF algorithm examines each data point in the dataset to calculate its LOF value. However, this approach can be slow and inefficient when applied to large datasets. To enhance efficiency, the LOF method can be improved by trimming the original data. Prior to applying the LOF algorithm, the original data is clustered using the DBSCAN (Density-Based Spatial Clustering of Applications with Noise) algorithm.

For the calculation of the LOF score, each cluster is assigned a center point. The average distance between the points in the cluster and its center is computed, representing the cluster's radius. For each point in the cluster, if its distance to the cluster's center is greater than or equal to the radius, it is added to the "Outlier Candidate Set." Consequently, the LOF score is only calculated for the data in the "Outlier Candidate Set," significantly reducing the computational load. An outlier is identified when its LOF score in the "Outlier Candidate Set" exceeds the mean plus four times the standard deviation (mean + 4SD).

\textbf{\textit{Missing Imputation}}
We adopted an Improved Random Forest imputation algorithm \cite{liu2020big} capable of effectively handling various types of missing data, including blocked missing data and completely missing horizontal or vertical data. The algorithm incorporates multiple imputation strategies, such as initializing imputation using linear interpolation for electricity supply/consumption data, prioritizing datasets with a small amount of missing data or less fluctuation, and leveraging the correlation between imputed variables and power data by concatenating them with other variables. Furthermore, in cases where an entire column of feature values is missing, the dataset is transformed into a horizontal row with a complete absence of data to overcome the limitation of constructing training sets for Random Forest.

Additionally, the algorithm employs Iterative Imputation, where each feature is modeled as a function of the other features, allowing sequential imputation using prior imputed values to predict subsequent features. This iterative process is repeated multiple times, leading to improved estimation of missing values across all features. To mitigate the potential issue of overfitting when using the Random Forest algorithm, we adopt cross-validation on a validation set. This helps us choose the effective iteration number and other hyperparameters, ensuring that the model's performance is evaluated on unseen data and preventing it from becoming too specialized to the training set.

\subsection{Prediction Model}

We present a novel approach, the GCN-LSTM encoder-decoder model, with a three-level attention mechanism for accurates line loss forecasting. The framework of our approach, as shown in Fig.~\ref{fig:framwork}, takes various inputs such as the topology graph of the distribution network, historical data with a time step of \begin{math}T\end{math} for meter reading, time-varying electrical parameters, and external influence information (historical weather data, transformer type and capacity, feeder length and type). These inputs are transformed into five matrices: an adjacency matrix, a line loss matrix, a time-varying electrical feature matrix, a weather index matrix, and a statical feature matrix. We combine the last four matrices into a mixed feature matrix.

Next, the mixed feature matrix is fed into a two-layer graph convolution network to capture the topological structure of the feeder and obtain the spatial features. The spatial features are then processed by an attention mechanism based on encoder-decoder to obtain distribution transformer-level importance scores and feature-level importance scores successively. These scores describe the significance of each distribution transformer and spatial feature, respectively.

The latent features, multiplied by the importance scores, are then input into a two-layer LSTM model to extract the corresponding temporal dependencies. An attention mechanism is employed to compute the significance score of the hidden state at distinct time steps. Finally, a fully connected layer network is used to convert the extracted features into the predicted values of line loss in the future at a specified time step.

\subsubsection{GCN Module}

The GCN is a powerful neural network that can directly act on arbitrarily structured graphs and utilize complex structural information. The distribution network is a graph model of distributed interconnected sensors, whose readings are modeled as time-dependent signals on the vertices \cite{bronstein2017geometric}. The power grid's topology has been explicitly used for approximating the power flow \cite{bolz2019power} and fault location \cite{chen2019fault}. Therefore, GCN is a suitable method for extracting features from the distribution network by utilizing their graph-based structure of interconnected distribution transformers. Our proposed model is based on the work of \cite{kipf2016semi}.

The graph convolution takes both the adjacency matrix \begin{math}A\end{math} and the graph-level features from the previous layer, \begin{math}H^{(l)}\in\end{math}\begin{math}\mathbb{R}^{N\times d_{l}} \end{math}, and outputs the graph-level features in the next layer, \begin{math}H^{(l+1)}\in\end{math}\begin{math}\mathbb{R}^{N\times d_{l+1}} \end{math}, where \begin{math}d_{l}\end{math} and \begin{math}d_{l+1}\end{math} are output feature dimension for layer \begin{math}l\end{math} and \begin{math}l+1\end{math}, respectively. This can be realized by equation ~\ref{eq:equation2}:
\begin{equation}
H^{(l+1)}=\sigma(AH^{(l)}W^{(l)})
\label{eq:equation2}
\end{equation}
where, \begin{math}H^{(0)}\end{math} is the mixed feature matrix \begin{math}C\end{math} from the result of the preprocessing of the input data, \begin{math}\sigma(.)\end{math} is a non-linear activation function like the ReLU, and the weight matrix \begin{math}W^{(l)}\in\end{math}\begin{math}\mathbb{R}^{d_{l}\times d_{l+1}} \end{math} is utilized to represent the trainable parameters of the neural network for layer \begin{math}l+1\end{math}.

Kipf and Welling identified two limitations associated with multiplication using the adjacency matrix \begin{math}A\end{math} in their paper. Firstly, this approach does not take into account the influence of a node on itself. Secondly, the adjacency matrix \begin{math}A\end{math} is not normalized, which can pose issues when attempting to extract graph features, as nodes with more neighbors may have a disproportionately larger impact. To overcome the initial constraint, the authors addressed the issue by introducing self-loops in the graph, which involved the addition of the identity matrix \begin{math}I\end{math} to \begin{math}A\end{math}. And they addressed the second limitation by using Symmetric normalized Laplacian, i.e., \begin{math}D^{-\frac{1}{2}}AD^{-\frac{1}{2}}\end{math}. By combining the two techniques, we can utilize the GCN model formulation introduced in \cite{kipf2016semi}, which can be expressed as:
\begin{equation}
H^{(l+1)}=\sigma(\widehat{D}^{-\frac{1}{2}}\widehat{A}\widehat{D}^{-\frac{1}{2}}H^{(l)}W^{(l)})
\end{equation}
Here, we use \begin{math}\widehat{A}=A+I\end{math}, where \begin{math}I\end{math} is the identity matrix, and \begin{math}\widehat{D}\end{math} is the diagonal node degree matrix of \begin{math}\widehat{A}\end{math}.

Our model performed best with a two-layer graph convolutional network (GCN), represented as:

\begin{equation}
GC(C, A){t}=ReLu(\widetilde{A}ReLu(\widetilde{A}CW{0})W_{1})
\end{equation}

Here, \begin{math}C\end{math} is the feature matrix obtained by merging input features, and \begin{math}\widetilde{A}\end{math} is the normalized adjacency matrix \begin{math}\widehat{A}\end{math} using the degree matrix \begin{math}\widehat{D}\end{math}. The weight matrix \begin{math}W_{0}\in\mathbb{R}^{(1+M+Q+K)\times h}\end{math} connects the input layer to the hidden layer, where \begin{math}(1+M+Q+K)\end{math} is the number of input features, and \begin{math}h\end{math} is the number of hidden units. The weight matrix \begin{math}W_{1}\in\mathbb{R}^{h\times O}\end{math} connects the hidden layer to the output layer, where \begin{math}O\end{math} is the number of output graph features, a hyperparameter. The output \begin{math}GC(C, A)_{t}\end{math} represents the graph features at time \begin{math}t\end{math}, with dimensions \begin{math}N\times O\end{math}, where \begin{math}N\end{math} is the number of nodes in the graph. The number of GCN units used in our model is the same as the length of the historical time series, denoted by \begin{math}T\end{math}, and the nonlinear activation function \begin{math}ReLU()\end{math} is applied.

\subsubsection{Attention Module}

The output of the GCN model is a 3-dimensional tensor \begin{math}G\in\end{math}\begin{math}\mathbb{R}^{T\times F\times N}\end{math}, where \begin{math}T\end{math} represents the duration of the historical time series, \begin{math}N\end{math} is the number of nodes (distribution transformers), and \begin{math}F\end{math} is the number of graph features. However, at every moment, not all nodes (distribution transformers), graph features, and previous time steps contribute equally to the total line loss. Therefore, we propose a three-level attention mechanism: distribution transformer-level attention, feature-level attention, and time-level attention, to focus on the most critical information responsible for loss prediction among all distribution transformers, graph features, and previous time steps, thus improving the prediction performance.

An attention mechanism allows a model to assign varying weights to different parts of the input, extracting more pertinent information and facilitating more precise decision-making without adding computational or storage costs. This can improve the accuracy of the model while also reducing its overall complexity. Attention mechanism was first applied in computer vision \cite{xu2015show}, and then was widely used in Seq2Seq model \cite{bahdanau2016end}, and generatesd many variants \cite{luong2015effective}\cite{shen2018disan}\cite{vaswani2017attention}. Inspired by the work done by an attention mechanism to obtain personalized biomarker scores \cite{beykikhoshk2020deeptriage}, we adopt a traditional Attention mechanism,  i.e., Soft Attention. Soft Attention is parameterization so it can be differentiable and can be embedded in the model for direct training.

After the GCN module,  the distribution transformer-level attention and feature-level attention are arranged sequentially to attend to the spatial features. But the time-level attention is located after the LSTM model to focus on each hidden state of LSTM and get the most crucial temporal information.
\begin{figure*}[ht]
\includegraphics[scale=0.22]{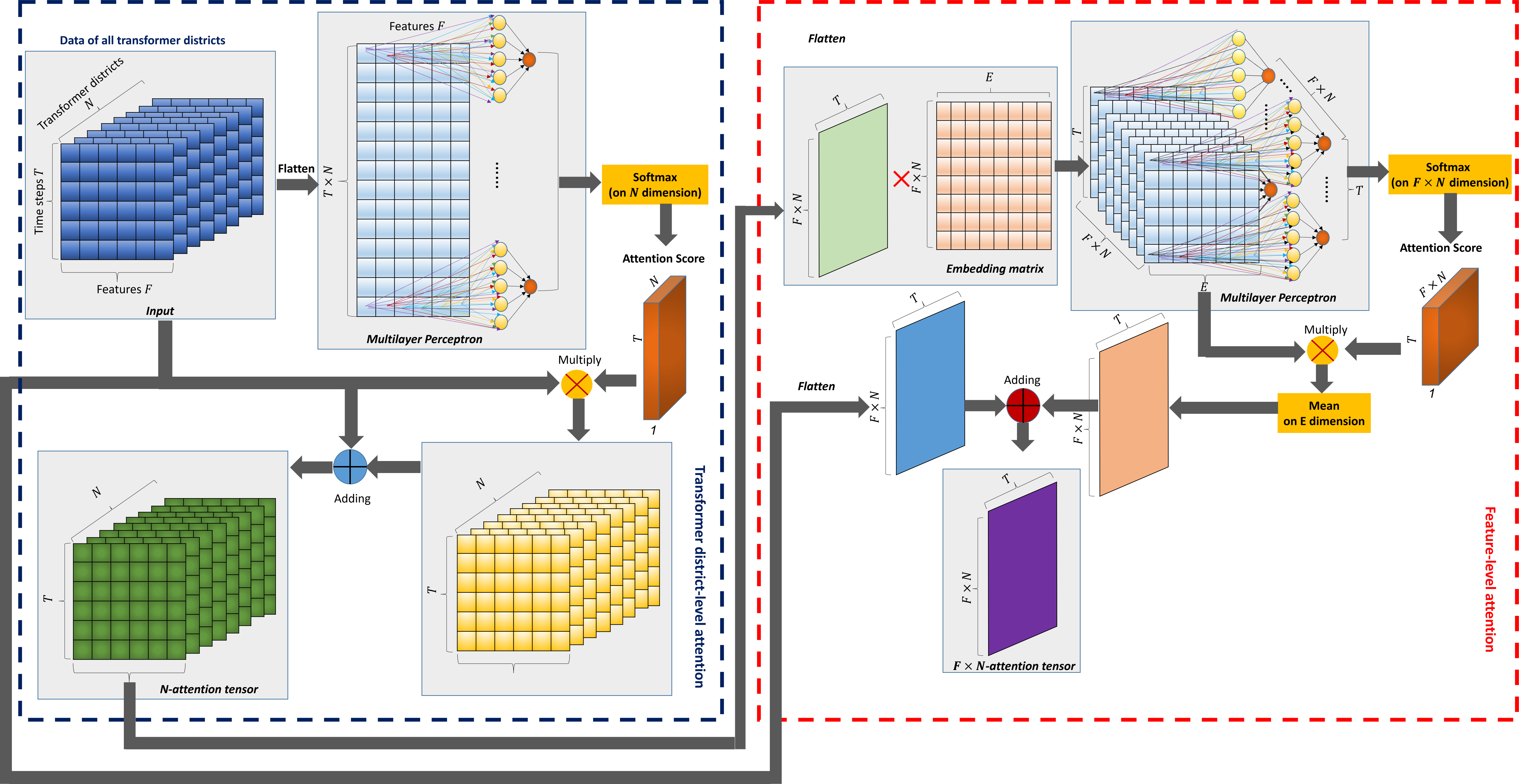}
\caption{Illustration of Distribution Transformer-level Attention (blue dotted frame) and Feature-level Attention mechanisms (red dotted frame). The former adjusts transformer importance at each time point by reshaping tensor $G(T, N, F)$ using a Multilayer Perceptron (Eq. \ref{eq:equation5}). The output tensor is then restored to $(T, N, 1)$, and softmax activation generates attention scores for each transformer (Eq. \ref{eq:equation6} and \ref{eq:equation7}). Then, the Feature-level attention mechanism flattens the N-attention tensor $(T, N, F)$ to $(T, N \times F)$, applies an embedding matrix, and performs a Multilayer Perceptron transformation (Eq. \ref{eq:equation8}). This process helps identify distinctive characteristics in the feature space. The resulting attention scores indicate the importance of each feature (Eq. \ref{eq:equation9}). Additionally, the $(T, N \times F, E)$ tensor is multiplied element-wise by the feature-level attention scores. The resulting tensor is then reduced along the E dimension by taking the mean, resulting in a $(T, N \times F)$ tensor. Finally, this tensor is added to the flattened original input using residual concatenation, allowing the model to retain important information and obtain the N × F-attention tensor $(T, N \times F)$.}
\label{fig:NFattention}
\end{figure*}

 \textbf{\textit{Distribution transformer-level attention:}} As shown in Fig.~\ref{fig:NFattention} blue dotted frame, the tensor \begin{math}G(T, N, F)\end{math} that comes from the GCN model is reshaped as \begin{math}(Z, F)\end{math}, where \begin{math}Z=T\times N\end{math}, and for every row \begin{math}Z_{i}\end{math}, a Multilayer Perceptron with one hidden layer and \begin{math}tanh\end{math}  nonlinearity is performed. The hidden size of the hidden layer is a hyperparameter tuned on the validation dataset. On the output layer, the output tensor's size is \begin{math}(Z, 1)\end{math}. This process can be expressed as:
\begin{equation}
\alpha_{i}= tanh(w_{1}(w_{0}Z_{i} + b_{0}) + b_{1})\quad and \quad i=1,2,...,T\times N
\label{eq:equation5}
\end{equation}
Then the output tensor is restore to size \begin{math}(T, N, 1)\end{math}, and use the softmax activation function across the \begin{math}N\end{math} (distribution transformers) dimension to ensure that the activations (\begin{math}\beta_{t}\end{math}, each corresponding entry of \begin{math}T\end{math}) are all between 0 and 1, and that they sum to 1, we call them attention score for each node (distribution transformer) at time  \begin{math}t\end{math}. It can be expressed as equation ~\ref{eq:equation6} ~\ref{eq:equation7}:
\begin{equation}
\begin{split}
\beta_{t}= [\beta_{t1},...,\beta_{tN}]
\label{eq:equation6}
\end{split}
\end{equation}
\begin{equation}
\begin{split}
\zeta_{t}= softmax(\beta_{t})=\frac{[\exp(\beta_{t1}),...,\exp(\beta_{tN})]}{\sum_{j=1}^{N}\exp(\beta_{tj})}
\end{split}
\label{eq:equation7}
\end{equation}
The attention score \begin{math}\zeta\in\end{math} \begin{math} \mathbb{R}^{T\times N\times 1} \end{math} is multiplied by the input, then the result is added to the input using residual concatenation to prevent information from being lost in the original sequence, and get the N-attention tensor \begin{math}(T, N, F)\end{math}.

\textbf{\textit{Feature-level attention:}} It follows the distribution transformer-level attention. As show in Fig.~\ref{fig:NFattention} red dotted frame, firstly, the  N-attention tensor \begin{math}(T, N, F)\end{math} is flatted as \begin{math}(T, N \times F)\end{math}, namely, there are \begin{math}N\end{math} distribution transformer, each distribution transformer has \begin{math}F\end{math} features, so each time sample has \begin{math} N \times F \end{math} features, and the feature value is expressed as \begin{math}x_{ji}\end{math}, where \begin{math}j \in [1 : T]\end{math}, \begin{math}i \in [1 : N \times F]\end{math}. Our aim is to identify distinctive characteristics at each time step through the utilization of feature embedding. This technique involves transforming the original high-dimensional feature space into a lower-dimensional space to facilitate efficient learning, as demonstratesd by \cite{golinko2017gfel}. An embedding matrix \begin{math}E_{(N \times F)\times d}\end{math} is defined, where each row represents a feature and each column represents a dimension of the embedding space. The d-dimensional embedding vector \begin{math}e_{i}=[e_{i1},\ldots, e_{id}]\end{math} corresponds to the i-th feature in the matrix, where \begin{math}i \in [1 : N \times F]\end{math}. To construct the embedding matrix, random vectors are used, which have been shown to perform similarly to end-to-end learning methods \cite{beykikhoshk2020deeptriage}. This same embedding matrix is applied to all time samples. Then, the flattened N-attention tensor is multiplied with the embedding matrix to produce a \begin{math}(T, N \times F, E)\end{math} tensor, which aims to capture the relationship between the feature values and the corresponding embedding vectors. Then as shown in equation ~\ref{eq:equation8}, a Multilayer Perceptron with one hidden layer and tanh nonlinearity is performed on each time point (\begin{math}T\end{math} dimension) and each feature (\begin{math}N \times F\end{math} dimension) simultaneously. On the output layer, the output tensor's size is \begin{math}(T, N \times F, 1)\end{math}. Next, as shown in equation~\ref{eq:equation9}, the softmax activation function across the \begin{math}N \times F\end{math} (features) dimension is used to get the attention score \begin{math} \xi_{j} \end{math} and indicate the different importance of each feature at each time sample. And the \begin{math}(T, N \times F, E)\end{math} tensor is multiplied by this attention score and take the mean on \begin{math}E\end{math} dimension to reduce to \begin{math}(T, N \times F)\end{math}. Finally, this result is added to the flattened original input using residual concatenation to avoid information loss and get the \begin{math} N \times F\end{math}-attention tensor \begin{math}(T, N \times F)\end{math}.
\begin{equation}
\begin{split}
\hat{\alpha}_{j}= tanh(w_{3}(w_{2}(x_{ji}\cdot e_{i} )+ b_{2}) + b_{3}) \\ and \quad j=1,2,...,T, \quad i=1,2,...,N\times F
\end{split}
\label{eq:equation8}
\end{equation}
\begin{equation}
\begin{split}
\xi_{j}= softmax(\hat{\alpha}_{j})=\frac{[\exp(\hat{\alpha}_{j1}),...,\exp(\hat{\alpha}_{j(N\times F)})]}{\sum_{i=1}^{N\times F}\exp(\hat{\alpha}_{ji})}
\end{split}
\label{eq:equation9}
\end{equation}

\begin{figure*}[ht]
\centering
\includegraphics[scale=0.37]{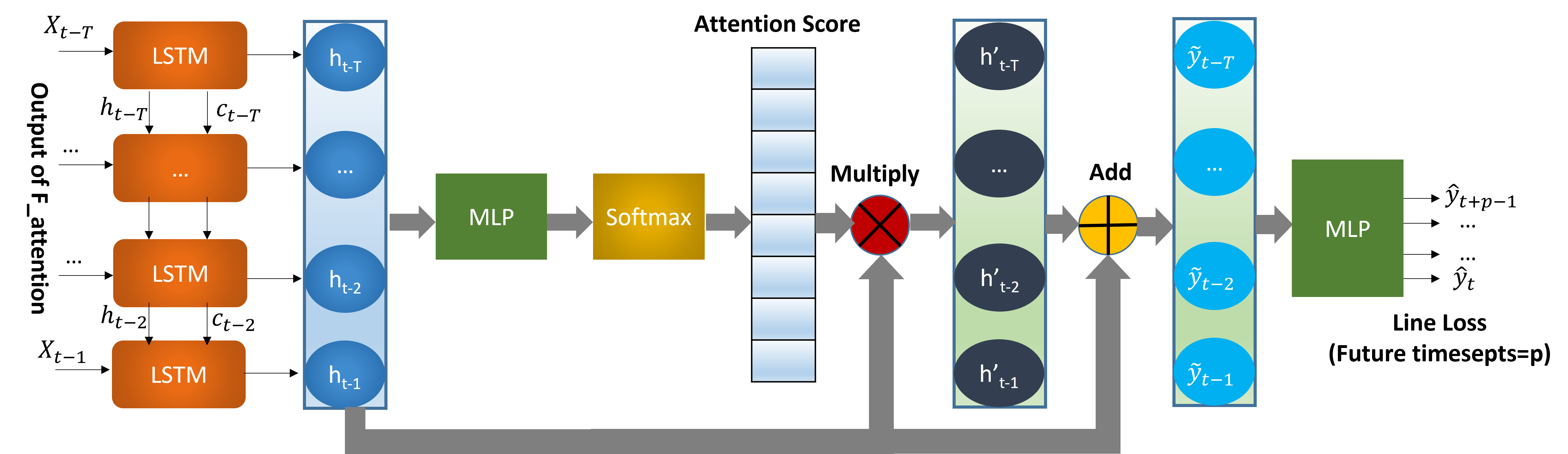}

\caption{Time-level Attention: Capturing Temporal Dependencies with LSTM and Attention Mechanism.  Unlike traditional approaches that solely utilize the last hidden state, our architecture attends to all hidden states, enabling the generation of a distinct vector representing the collective information. This is achieved by assigning different attention scores to each hidden state. The attention scores are computed through a Multilayer Perceptron transformation (Eq. ~\ref{eq:equation16}) followed by the application of the softmax activation function (Eq. ~\ref{eq:equation17}). The resulting attention-weighted hidden states are then aggregated to obtain the context vector (Eq. ~\ref{eq:equation18}), which captures the overall variation in the power grid.}
\label{fig:Tattention}
\end{figure*}
\textbf{\textit{Time-level attention:}} At this level, we use LSTM model to capture the temporary dependency of the output of the previous feature-level attention, LSTM is an improvement on the hidden layer of Recurrent Neural Network (RNN). It can be used to solve long-term dependence problems and can remember long-term information \cite{hochreiter1997long}. The operation equation is given in equation ~\ref{eq:equation10} -~\ref{eq:equation15}\cite{graves2006connectionist}. Besides, the ``attention" will also be used to focus more on relevant information of the time sequence, thus improving the prediction performance.
\begin{equation}
f_t=\sigma(w_p\cdot [h_{t-1},x_t ]+b_f)
\label{eq:equation10}
\end{equation}
\begin{equation}
i_t=\sigma(w_i\cdot [h_{t-1},x_t ]+b_i)
\end{equation}
\begin{equation}
g_t=tanh(w_g\cdot[h_{t-1},x_t ]+b_g)
\end{equation}
\begin{equation}
c_t=f_t\ast c_{t-1}+i_t\ast g_t
\end{equation}
\begin{equation}
o_t=\sigma(w_o\cdot [h_(t-1),x_t ]+b_o)
\end{equation}
\begin{equation}
h_t=o_t\ast tanh(c_t)
\label{eq:equation15}
\end{equation}
\

As shown in Fig.~\ref{fig:Tattention}, The output of the previous feature-level attention, \begin{math}X_{t-1},...X_{t-T}\end{math} denote \begin{math}T\end{math} time steps sequence before the current time step, each time step \begin{math}X_{i}=[x_{i1},...,x_{iN \times F }]\end{math}, where \begin{math}N \times F\end{math} is the dimension of the feature space. We use an architecture that attend to all hidden states instead of only using the last hidden state of LSTM model as proxy. It produces a distinct vector representing all hidden states by giving different attention scores to different hidden states. First, a Multilayer Perceptron with one hidden layer and tanh nonlinearity is performed on the hidden states of LSTM \begin{math}H=[h_{t-1},...,h_{t-T}]\end{math}, then, the attention score of each hidden state are calculated by a softmax activation function. After that, by adding the product of attention score and input hidden states to the input hidden states, we get the context vector \begin{math}\widetilde{y_{i}}\end{math} that covers overall power grid variation information is shown in equation ~\ref{eq:equation18}.

\begin{equation}
\begin{split}
\eta_{i}= tanh(w_{5}(w_{4}H+ b_{4}) + b_{5}) \\ and \quad i=t-1,t-2,...,t-T
\end{split}
\label{eq:equation16}
\end{equation}
\begin{equation}
\begin{split}
\nu_{i}= softmax(\eta_{i})=\frac{\exp(\eta_{i})}{\sum_{k=t-1}^{t-T}\exp(\eta_{k})}
\end{split}
\label{eq:equation17}
\end{equation}
\begin{equation}
\begin{split}
\widetilde{y_{i}}=\nu_{i}\times h_{i}+h_{i}
\end{split}
\label{eq:equation18}
\end{equation}

Finally, the prediction results for future line loss rates over \begin{math}p\end{math} time steps are obtained by passing the output of the LSTM layer, \begin{math}\widetilde{y_{i}}\end{math}, through a fully connected layer. This can be represented as:
\begin{equation}
\hat{y} = \text{ReLU}(W_{FC} \cdot \widetilde{y_{i}} + b_{FC})
\label{eq:fully_connected}
\end{equation}

where \begin{math}\hat{y}\end{math} represents the predicted line loss rates, \begin{math}W_{FC}\end{math} and \begin{math}b_{FC}\end{math} are the weight matrix and bias vector of the fully connected layer, and $ReLU$ denotes the Rectified Linear Unit activation function.

\subsection{Training Procedure}
The learning process of the attention-GCN-LSTM model involves end-to-end training, where all components are jointly optimized to improve the model's performance. The mean squared error (MSE) is commonly adopted as a loss function to evaluate the prediction accuracy of the model. To ensure the stability and convergence of the attention-GCN-LSTM model, we monitor its performance on a validation set during training. This helps prevent overfitting and allows for the adjustment of hyperparameters, including the learning rate and regularization strength, to find the optimal values. We utilize the widely adopted Adam optimizer, known for its fast convergence, to optimize the loss function. Additionally, gradient clipping is applied during training to prevent the gradient explosion. To avoid unnecessary training iterations, we implement an early stopping mechanism that halts the training process if the mean squared error (MSE) on the validation set remains unchanged for 20 consecutive iterations.

\section{EXPERIMENTS}

\textbf{\textit{Data Discription}}
In this study, we collected a comprehensive dataset from 44 distribution transformers located in Lingling, Hunan province, China. The dataset included various electrical measurements such as three-phase current, voltage, active power, and reactive power, recorded every 15 minutes. We also incorporated substation gateway data collected every half an hour and daily electric supply and consumption data for missing data imputation. The dataset covered a significant time period from January 1, 2017, to July 31, 2018. From the dataset, we derived important parameters such as load rates, power factor, and three-phase imbalance degree, providing insights into energy usage efficiency and power system stability. Using the PowerFactory software, we created a simulation model of the 10KV distribution network to calculate power flow and determine the historical overall feeder loss rate. To analyze the relationship between weather conditions and electrical parameters, we integrated hourly weather data obtained from the Meteostat Python library. To ensure compatibility with the hourly sampling frequency of the weather data, we performed a resampling process. This involved selecting the data points recorded at 15-minute intervals and retaining only the data points corresponding to each one-hour interval. By aligning our electrical data with the hourly intervals of the weather data, we enabled accurate analysis and meaningful comparisons between the two datasets. This resampling process allowed us to unify the time granularity and ensure consistency between the electrical features and the weather data. The data were split into Train, Validation, and Test sets using an 8:1:1 ratio, respectively.

\begin{table*}[ht]
\centering
\caption{The key hyperparameter values for each baseline were optimized and fine-tuned according to the respective baseline papers, taking into consideration the specific recommendations and guidelines provided in each paper.}
\begin{tabular}{cc}
   \toprule
   Model & Design   \\
   \midrule
   ARIMA & p=2, d=1, q=1  \\
   BP & epoch= 200,batch\_size=64,lr=1e-4,weight\_decay=5e-4, dropout=0.05,hidden-layers=2 \\
   SVR & n\_components=50, ker  = 'rbf', C = 2, epsi = 0.001, par  = 0.8, tol = 1e-10  \\
   RF & n\_components=50, n\_estimators=100, max\_depth=30, random\_state=0 \\
   GBDT & lr=1e-4, n\_estimators=2000, max\_depth=15, max\_features='sqrt', min\_samples\_leaf=10, min\_samples\_split=10,loss='ls', random\_state =42 \\
   LSTM& epoch= 200,batch\_size=64,lr=1e-4,weight\_decay=5e-4, dropout=0.05,hidden-layers=2,hidden\_size=256 \\
   Bi-LSTM& epoch= 200,batch\_size=64,lr=1e-3,hidden-layers=2,hidden\_size=256 \\
   TCN-LSTM& epoch= 200,batch\_size=64,lr=1e-4,num\_layers=2,weight\_decay=5e-4, dropout=0.05,hidden\_size=[128,64] kernel\_size=3\\
   DBN& epoch= 200,batch\_size=64,lr=1e-4,weight\_decay=5e-4, dropout=0.05,hidden-layers=2,hidden\_size=[128,64]\\
   PE-CNN& epoch= 200,batch\_size=64,lr=1e-4,weight\_decay=5e-4, dropout=0.05,num\_filters = [16,64,128,256],kernel\_sizes = [5,3,3,3],max\_seq\_len=100\\
   \bottomrule

\end{tabular}
\label{Tab:baseline}
\end{table*}

\textbf{\textit{Implementation details}}
The model parameters are adjusted using the validation set. We employed a combination of grid search and random search to identify the optimal parameter configuration for our proposed model. We adopt a sliding window with size of 100 for all baseline to obtain a set of sub-series. To optimize our model, we utilize the ADAM optimizer \cite{kingma2015variational} with an initial learning rates of \begin{math}10^{-4}\end{math}. Our GCN architecture comprises two layers with output dimensions of 256 and 16, respectively. We set the layer of LSTM as 2,and the size of hidden state of LSTM as 256. The training process is early stopped within 100  with the batch size of 32. All the experiments are implemented in Pytorch \cite{paszke2019pytorch} and conducted on a single NVIDIA TITAN RTX 12GB GPUs. The parameters of the baseline models are shown in Table~\ref{Tab:baseline}.

\textbf{\textit{Baselines}}
We have conducted a comparative analysis of our model against several baselines, including: (1) Auto-Regressive Integratesd Moving Average (ARIMA)\cite{badrinath2015arima}, (2) BP neural network \cite{yitao201910}, (3) Support Vector Regression (SVR)\cite{min2015application}, (4) Random Forest (RF) \cite{wang2017line}, (5) Gradient Boosted Decision Trees (GBDT)\cite{yao2019research}, (6) Long short-term memory (LSTM)\cite{liu2023analysis}, (7) bidirectional LSTM (Bi-LSTM) \cite{tulensalo2020lstm}, (8) Temporal Convolutional Network LSTM (TCN-LSTM) \cite{bi2021hybrid}, (9) Deep belief network (DBN) \cite{wang2019adaptive}, (10) Position Encoding-Convolutional Neural Networks (PE-CNN) \cite{jin2022position}.

\textbf{\textit{Evaluation metrics}}
In order to evaluate the predictive ability of the model, we utilize Root Mean Squared Error (RMSE), Mean Absolute Error (MAE), and Coefficient of Determination (\begin{math}R^{2}\end{math}). Smaller values of RMSE and MAE indicate a smaller error, which translates to higher precision in prediction. On the other hand, \begin{math}R^{2}\end{math} is a widely used metric to assess the strengths and weaknesses of regression models, where a value closer to 1 reflects a better fit of the regression equation.

\begin{figure*}[ht]
  \centering
  \begin{subfigure}[b]{0.47\textwidth}
    \includegraphics[width=\textwidth]{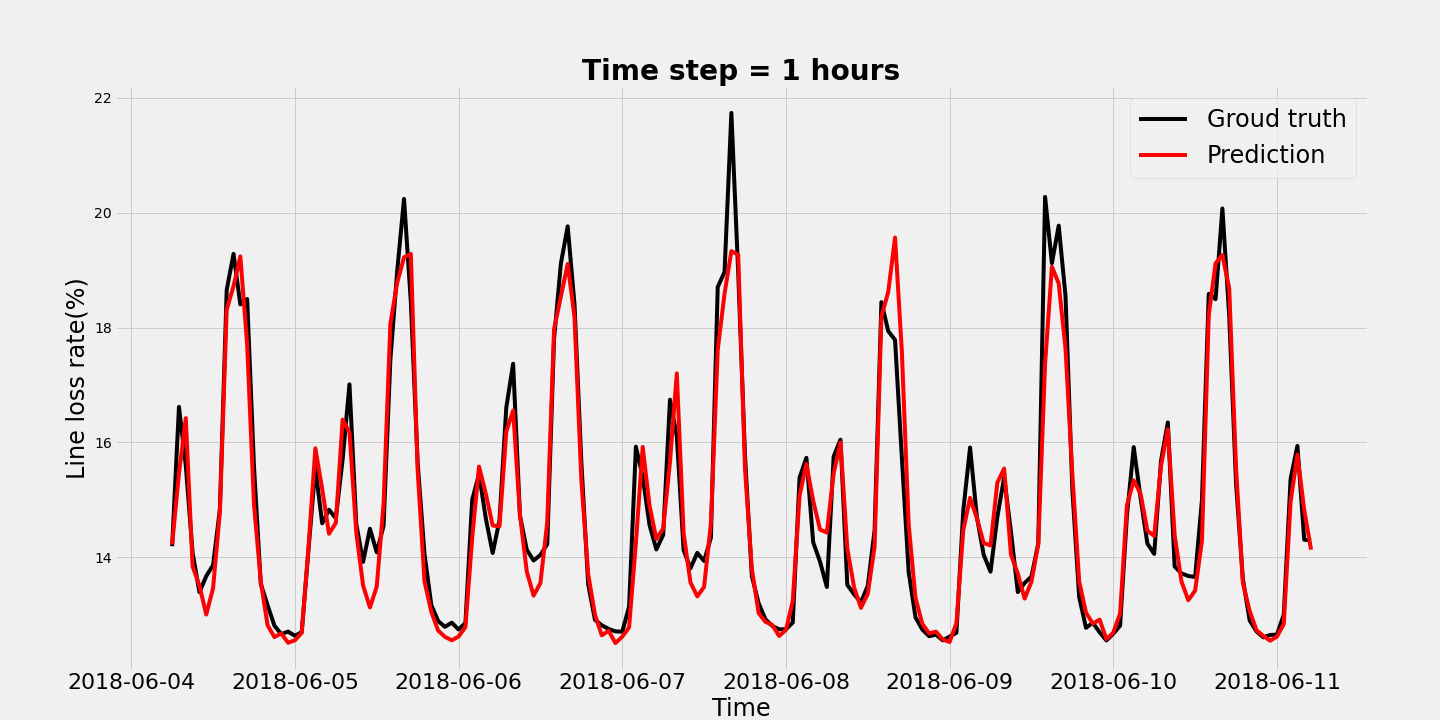}

  \hspace*{-0.6in}
  \end{subfigure}
  \begin{subfigure}[b]{0.47\textwidth}
    \includegraphics[width=\textwidth]{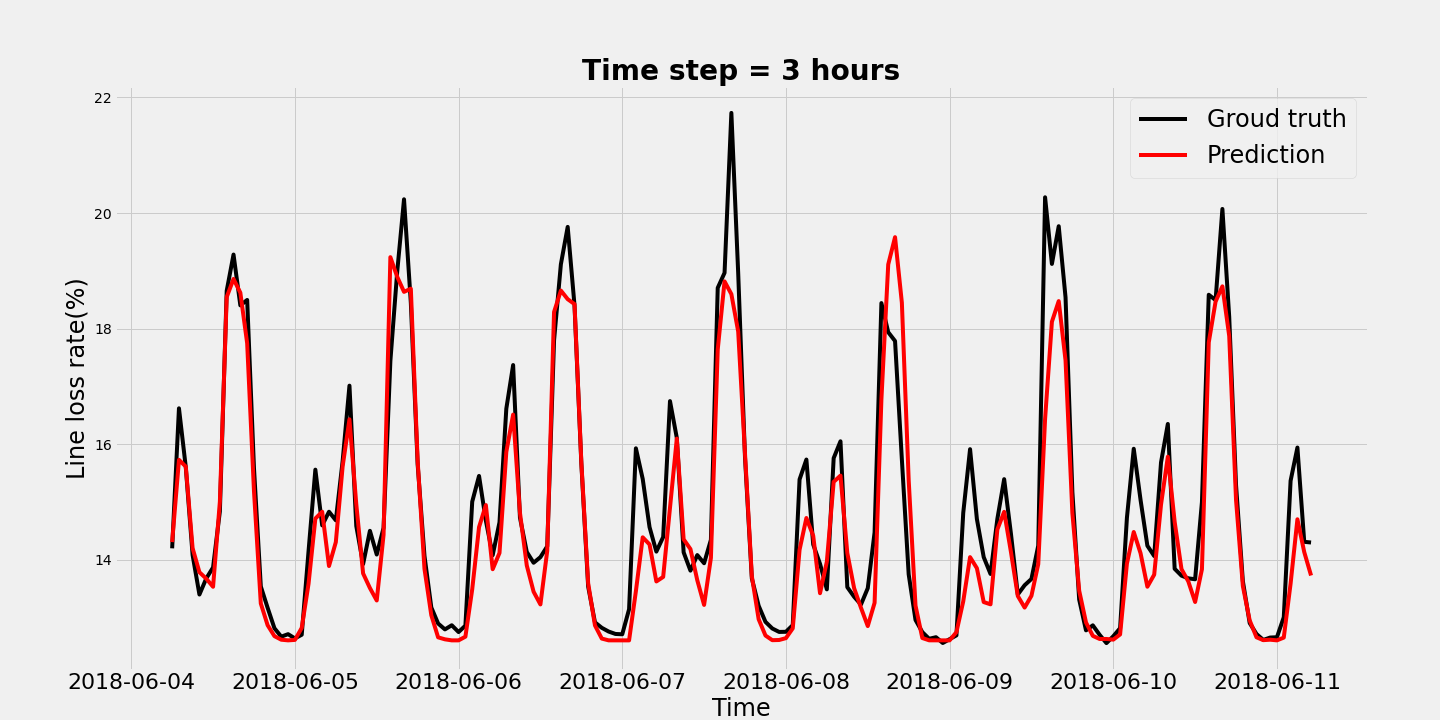}

  \hspace*{-0.6in}
  \end{subfigure}
  \begin{subfigure}[b]{0.47\textwidth}
    \includegraphics[width=\textwidth]{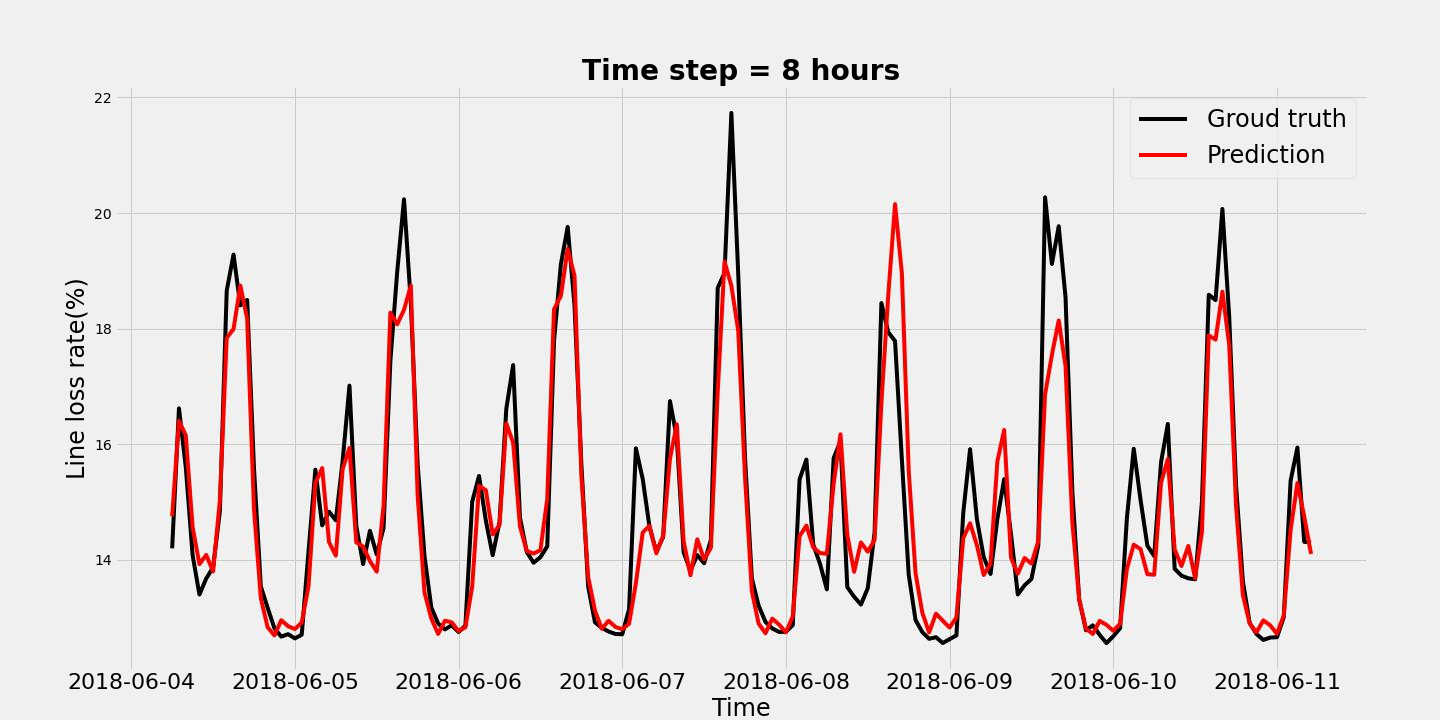}

  \hspace*{-0.6in}
  \end{subfigure}
  \begin{subfigure}[b]{0.47\textwidth}
    \includegraphics[width=\textwidth]{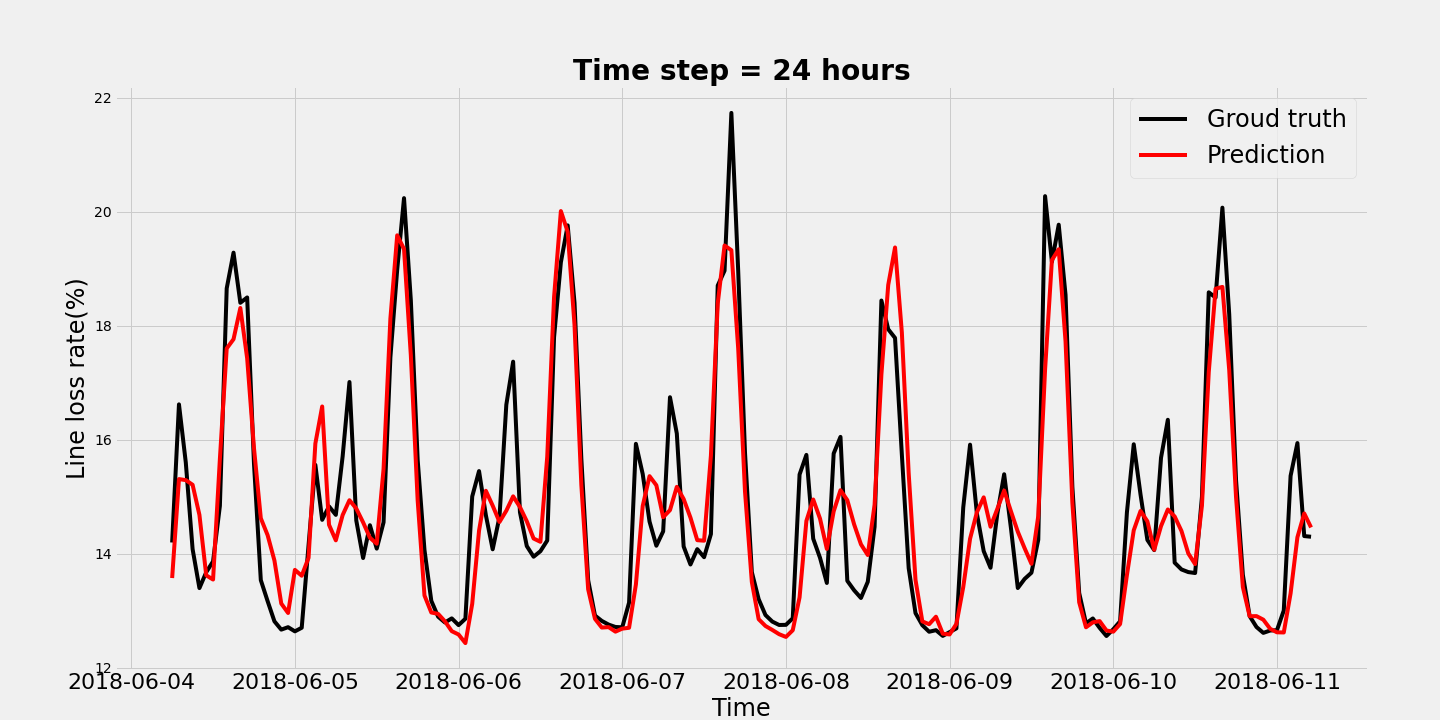}

  \hspace*{-0.6in}
  \end{subfigure}
  \begin{subfigure}[b]{0.47\textwidth}
    \includegraphics[width=\textwidth]{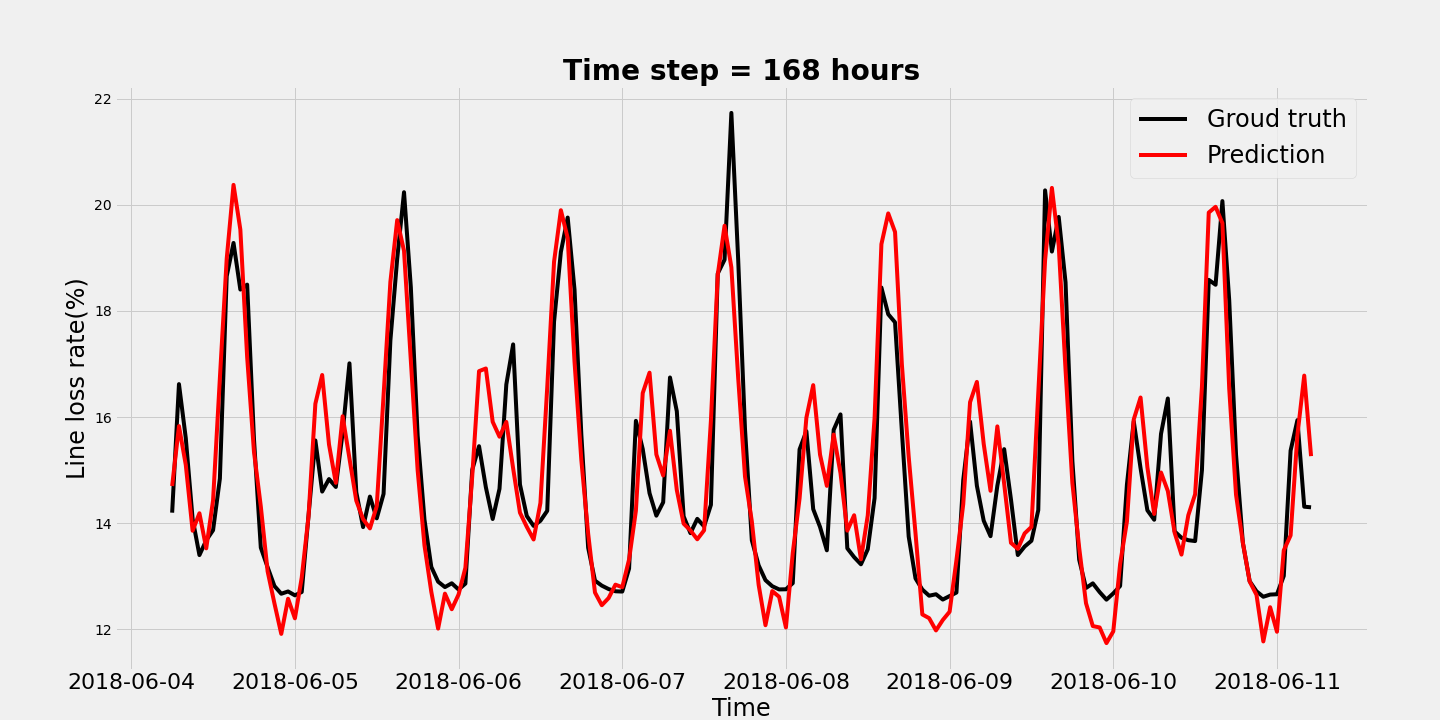}

  \hspace*{-0.6in}
  \end{subfigure}

  \caption{the visualization results of the overall line loss of the feeder under prediction horizons of 1 hour, 3 hours, 8 hours, and 168 hours from Jun 4 to Jun 11, 2018.}
\label{fig:results}
\end{figure*}

\begin{table*}[htb]
\centering
\caption{Performance comparison of our Attention-GCN-LSTM model with other baselines in line loss rates prediction at different time horizons. The metrics include Root Mean Squared Error (RMSE), Mean Absolute Error (MAE), and R2 score. The top-performing outcomes are denoted in bold.}
\begin{tabular}{|l|c|lcllllllllc|}
\hline
\multicolumn{1}{|c|}{\multirow{2}{*}{\textbf{T}}}      & \multicolumn{1}{l|}{\multirow{2}{*}{\textbf{Metric}}} & \multicolumn{11}{c|}{\textbf{Algorithm}}                                                                                                                                                                                                                                       \\ \cline{3-13}
\multicolumn{1}{|c|}{}                                 & \multicolumn{1}{l|}{}                                 & \multicolumn{1}{l|}{\textbf{ARIMA}} & \multicolumn{1}{c|}{\textbf{RF}} & \multicolumn{1}{l|}{\textbf{GBDT}} & \multicolumn{1}{l|}{\textbf{SVR}}    & \multicolumn{1}{l|}{\textbf{BP}} & \multicolumn{1}{l|}{\textbf{LSTM}} &\multicolumn{1}{l|}{\textbf{Bi-LSTM}} & \multicolumn{1}{l|}{\textbf{TCN-LSTM}} &  \multicolumn{1}{l|}{\textbf{DBN}}   &  \multicolumn{1}{l|}{\textbf{PE-CNN}}& \multicolumn{1}{l|}{\textbf{Ours}} \\ \hline
\multicolumn{1}{|c|}{\multirow{3}{*}{\textbf{1 hour}}} & \textbf{RMSE}                                         & \multicolumn{1}{l|}{1.2162}         & \multicolumn{1}{c|}{0.9468}      & \multicolumn{1}{l|}{1.1304}        & \multicolumn{1}{l|}{1.4043}          & \multicolumn{1}{l|}{0.9871}      & \multicolumn{1}{l|}{0.9359} & \multicolumn{1}{l|}{0.9452}       &  \multicolumn{1}{l|}{0.8352}       & \multicolumn{1}{l|}{0.8068}       & \multicolumn{1}{l|}{0.8308}       &\textbf{0.5795}                                  \\
\multicolumn{1}{|c|}{}                                 & \textbf{MAE}                                          & \multicolumn{1}{l|}{0.9014}         & \multicolumn{1}{c|}{0.7079}      & \multicolumn{1}{l|}{0.8632}        & \multicolumn{1}{l|}{1.0940}          & \multicolumn{1}{l|}{0.7558}      & \multicolumn{1}{l|}{0.7115}     & \multicolumn{1}{l|}{0.6940}    &  \multicolumn{1}{l|}{0.5838}       & \multicolumn{1}{l|}{0.5902}       & \multicolumn{1}{l|}{0.6232}       & \textbf{0.3947}                                  \\
\multicolumn{1}{|c|}{}                                 & \textbf{ \begin{math}\mathbf{R^{2}}\end{math}}                                           & \multicolumn{1}{l|}{0.7279}         & \multicolumn{1}{c|}{0.8342}      & \multicolumn{1}{l|}{0.7637}        & \multicolumn{1}{l|}{0.6353}          & \multicolumn{1}{l|}{0.8198}      & \multicolumn{1}{l|}{0.8362}    & \multicolumn{1}{l|}{0.8330}     &  \multicolumn{1}{l|}{0.8696}       & \multicolumn{1}{l|}{0.8783}       & \multicolumn{1}{l|}{0.8709}       & \textbf{0.9241}                                  \\ \hline
\multirow{3}{*}{\textbf{3 hours}}                      & \textbf{RMSE}                                         & \multicolumn{1}{l|}{1.7574}         & \multicolumn{1}{c|}{0.9989}      & \multicolumn{1}{l|}{1.2817}        & \multicolumn{1}{l|}{1.1334}          & \multicolumn{1}{l|}{1.3080}      & \multicolumn{1}{l|}{1.2886}      & \multicolumn{1}{l|}{1.2918}   &  \multicolumn{1}{l|}{1.1015}       & \multicolumn{1}{l|}{0.8829}       & \multicolumn{1}{l|}{0.8845}       & \textbf{0.7490}                                  \\
                                                       & \textbf{MAE}                                          & \multicolumn{1}{l|}{1.3518}         & \multicolumn{1}{c|}{0.7499}      & \multicolumn{1}{l|}{0.9644}        & \multicolumn{1}{l|}{0.8173}          & \multicolumn{1}{l|}{1.0328}      & \multicolumn{1}{l|}{0.9751}   & \multicolumn{1}{l|}{1.0065}      &  \multicolumn{1}{l|}{0.8405}       & \multicolumn{1}{l|}{0.6501}       & \multicolumn{1}{l|}{0.6624}       & \textbf{0.5024}                                  \\
                                                       & \textbf{\begin{math}\mathbf{R^{2}}\end{math}}                                           & \multicolumn{1}{l|}{0.4323}         & \multicolumn{1}{c|}{0.8151}      & \multicolumn{1}{l|}{0.6957}        & \multicolumn{1}{l|}{0.7621}          & \multicolumn{1}{l|}{0.6831}      & \multicolumn{1}{l|}{0.6895}        & \multicolumn{1}{l|}{0.6880}  &  \multicolumn{1}{l|}{0.7731}       & \multicolumn{1}{l|}{0.8542}       & \multicolumn{1}{l|}{0.8537}       & \textbf{0.8734}                                  \\ \hline

\multirow{3}{*}{\textbf{8 hours}}                      & \textbf{RMSE}                                         & \multicolumn{1}{l|}{2.0341}         & \multicolumn{1}{c|}{1.3194}      & \multicolumn{1}{l|}{1.2504}        & \multicolumn{1}{l|}{1.0444}          & \multicolumn{1}{l|}{1.2056}      & \multicolumn{1}{l|}{1.3289}     & \multicolumn{1}{l|}{1.3984}    &  \multicolumn{1}{l|}{1.3213}       & \multicolumn{1}{l|}{1.3677}       & \multicolumn{1}{l|}{0.9256}       & \textbf{0.7923}                                  \\
                                                       & \textbf{MAE}                                          & \multicolumn{1}{l|}{1.5641}         & \multicolumn{1}{c|}{0.8691}      & \multicolumn{1}{l|}{0.9640}        & \multicolumn{1}{l|}{0.7119} & \multicolumn{1}{l|}{0.9371}      & \multicolumn{1}{l|}{0.9982}    & \multicolumn{1}{l|}{1.0639}     &  \multicolumn{1}{l|}{0.9955}       & \multicolumn{1}{l|}{1.0159}       & \multicolumn{1}{l|}{0.6955}       & \textbf{0.5292}                                           \\
                                                       & \textbf{\begin{math}\mathbf{R^{2}}\end{math}}                                           & \multicolumn{1}{l|}{0.2360}         & \multicolumn{1}{c|}{0.6763}      & \multicolumn{1}{l|}{0.7055}        & \multicolumn{1}{l|}{0.7971}          & \multicolumn{1}{l|}{0.7297}      & \multicolumn{1}{l|}{0.6698}        & \multicolumn{1}{l|}{0.6345}  &  \multicolumn{1}{l|}{0.6737}       & \multicolumn{1}{l|}{0.6503}       & \multicolumn{1}{l|}{0.8399}       & \textbf{0.8584}                                  \\ \hline
\multirow{3}{*}{\textbf{24 hours}}                      & \textbf{RMSE}                                         & \multicolumn{1}{l|}{2.1801}         & \multicolumn{1}{c|}{1.4377}      & \multicolumn{1}{l|}{1.2616}        & \multicolumn{1}{l|}{1.0362}          & \multicolumn{1}{l|}{1.1570}      & \multicolumn{1}{l|}{1.3524}    & \multicolumn{1}{l|}{1.4641}    &  \multicolumn{1}{l|}{2.2447}       & \multicolumn{1}{l|}{1.8410}       & \multicolumn{1}{l|}{1.1087}        & \textbf{0.9169}                                  \\
                                                       & \textbf{MAE}                                          & \multicolumn{1}{l|}{1.6779}         & \multicolumn{1}{c|}{0.9236}      & \multicolumn{1}{l|}{0.9588}        & \multicolumn{1}{l|}{0.7154}          & \multicolumn{1}{l|}{0.8951}      & \multicolumn{1}{l|}{1.0101}    & \multicolumn{1}{l|}{1.1774}     &  \multicolumn{1}{l|}{1.7090}       & \multicolumn{1}{l|}{1.4887}       & \multicolumn{1}{l|}{0.8574}       & \textbf{0.6748}                                  \\
                                                       & \textbf{\begin{math}\mathbf{R^{2}}\end{math}}                                           & \multicolumn{1}{l|}{0.1228}         & \multicolumn{1}{c|}{0.6163}      & \multicolumn{1}{l|}{0.7045}        & \multicolumn{1}{l|}{0.8007}          & \multicolumn{1}{l|}{0.7507}      & \multicolumn{1}{l|}{0.6579}        & \multicolumn{1}{l|}{0.6009}  &  \multicolumn{1}{l|}{0.0619}       & \multicolumn{1}{l|}{0.3690}       & \multicolumn{1}{l|}{0.7712}       & \textbf{0.8101}                                  \\ \hline
\multirow{3}{*}{\textbf{168 hours}}                      & \textbf{RMSE}                                         & \multicolumn{1}{l|}{2.3142}         & \multicolumn{1}{c|}{1.6907}      & \multicolumn{1}{l|}{1.4299}        & \multicolumn{1}{l|}{1.4357}          & \multicolumn{1}{l|}{1.3247}      & \multicolumn{1}{l|}{2.6213}       & \multicolumn{1}{l|}{1.9260}   &  \multicolumn{1}{l|}{2.4603}       & \multicolumn{1}{l|}{2.3598}       & \multicolumn{1}{l|}{1.3270} & \textbf{1.0119}                                  \\
                                                       & \textbf{MAE}                                          & \multicolumn{1}{l|}{1.7706}         & \multicolumn{1}{c|}{1.0484}      & \multicolumn{1}{l|}{1.0751}        & \multicolumn{1}{l|}{1.0358}          & \multicolumn{1}{l|}{1.0315}      & \multicolumn{1}{l|}{2.2929}     & \multicolumn{1}{l|}{1.5324}    &  \multicolumn{1}{l|}{1.9754}       & \multicolumn{1}{l|}{1.8427}       & \multicolumn{1}{l|}{1.0193} & \textbf{0.7859}                                  \\
                                                       & \textbf{\begin{math}\mathbf{R^{2}}\end{math}}                                           & \multicolumn{1}{l|}{0.0013}         & \multicolumn{1}{c|}{0.4648}      & \multicolumn{1}{l|}{0.6171}        & \multicolumn{1}{l|}{0.6140}          & \multicolumn{1}{l|}{0.6714}      & \multicolumn{1}{l|}{-0.2869}        & \multicolumn{1}{l|}{-0.3053}  &  \multicolumn{1}{l|}{-0.1337}       & \multicolumn{1}{l|}{-0.0429}       & \multicolumn{1}{l|}{0.6702} & \textbf{0.7687}                                  \\ \hline
\end{tabular}
\label{Tab:results}
\end{table*}

\begin{figure*}[ht]
  \centering
  \begin{subfigure}[b]{0.31\textwidth}
    \includegraphics[width=\textwidth]{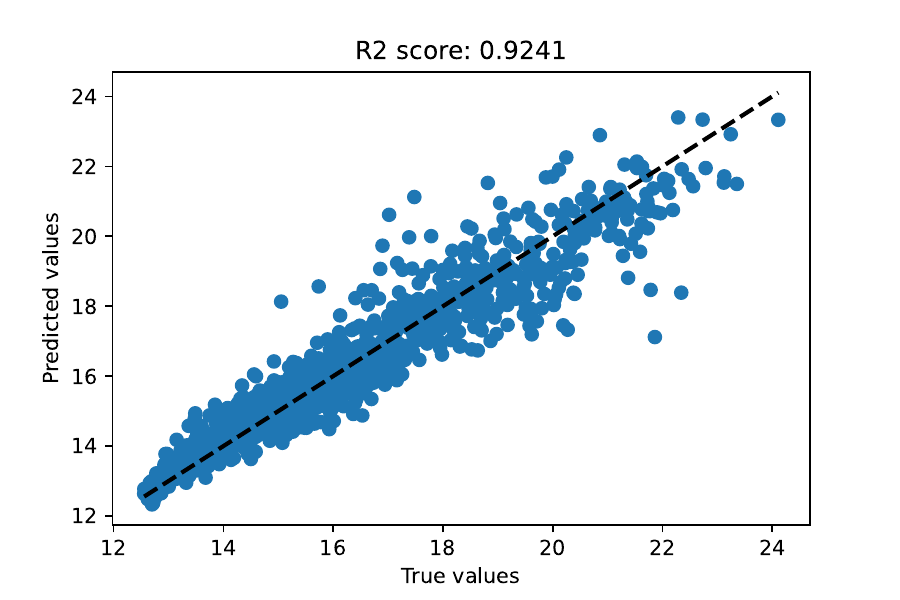}
    \caption{Attention-GCN-LSTM}
  \end{subfigure}
  \begin{subfigure}[b]{0.31\textwidth}
    \includegraphics[width=\textwidth]{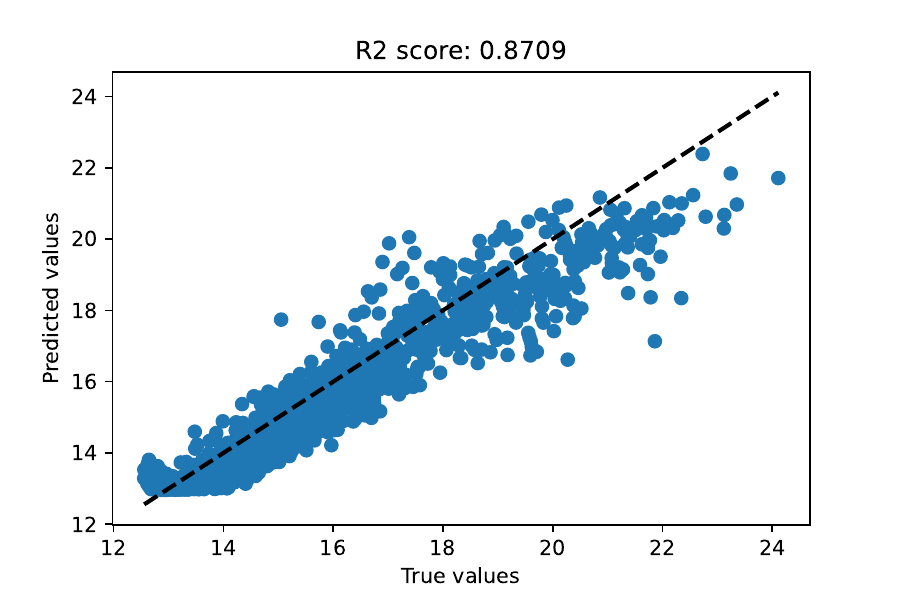}
    \caption{PE-CNN}
  \end{subfigure}
  \begin{subfigure}[b]{0.31\textwidth}
    \includegraphics[width=\textwidth]{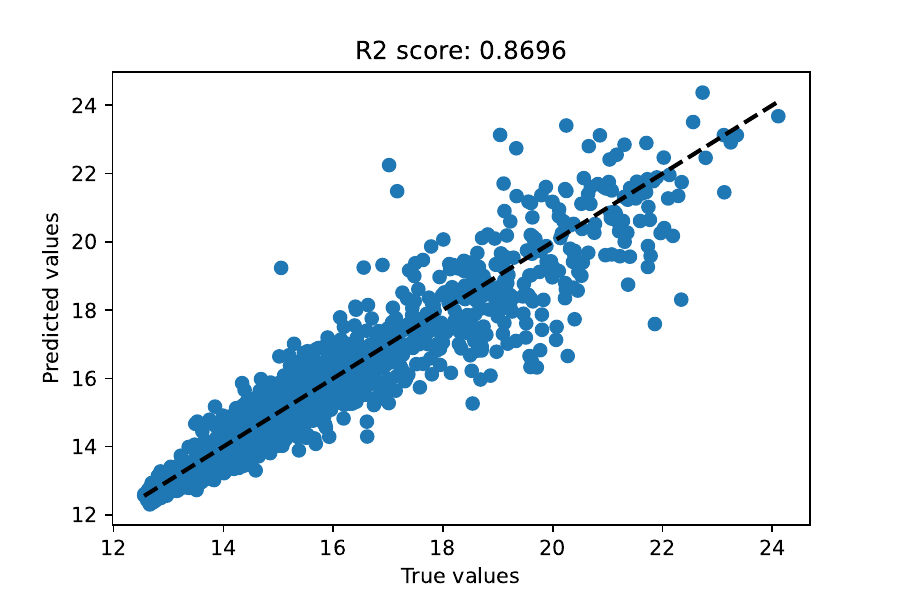}
    \caption{TCN-LSTM}
  \end{subfigure}
  \begin{subfigure}[b]{0.31\textwidth}
    \includegraphics[width=\textwidth]{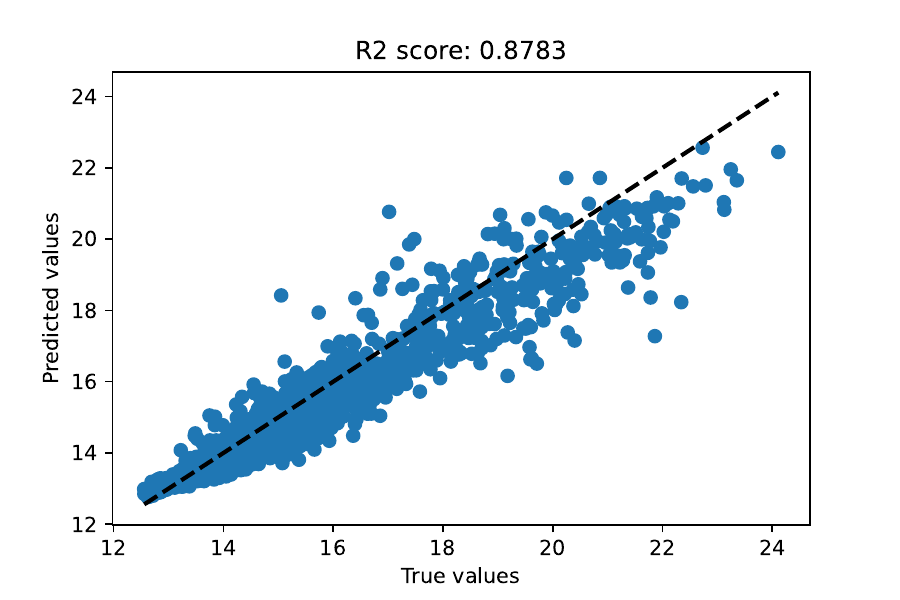}
    \caption{DBN}
  \end{subfigure}
  \begin{subfigure}[b]{0.31\textwidth}
    \includegraphics[width=\textwidth]{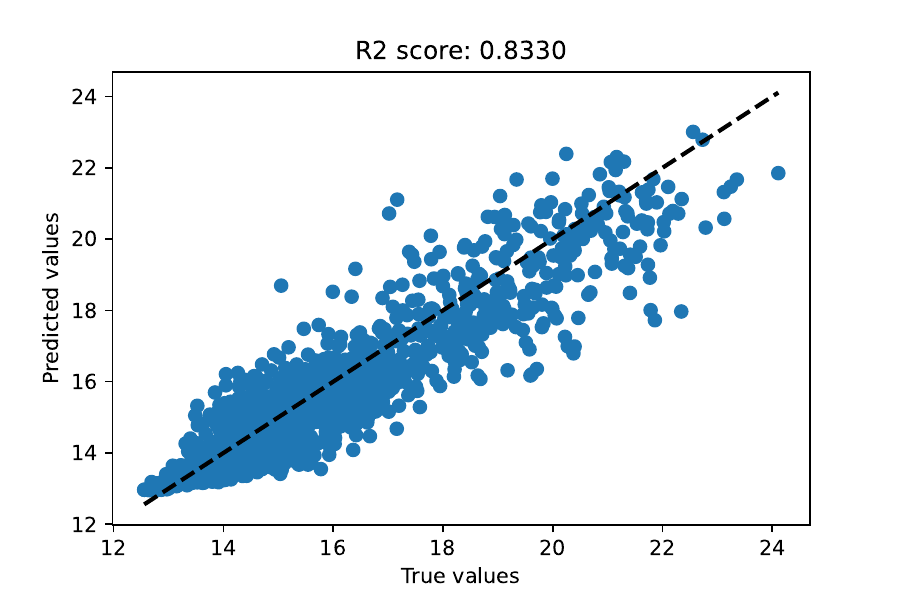}
    \caption{Bi-LSTM}
  \end{subfigure}
  \begin{subfigure}[b]{0.31\textwidth}
    \includegraphics[width=\textwidth]{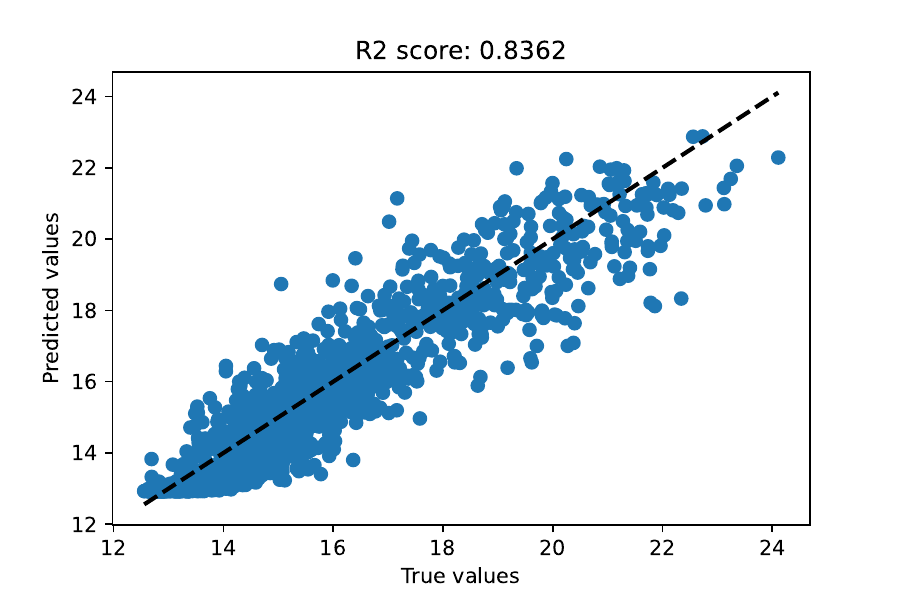}
    \caption{LSTM}
  \end{subfigure}
  \begin{subfigure}[b]{0.31\textwidth}
    \includegraphics[width=\textwidth]{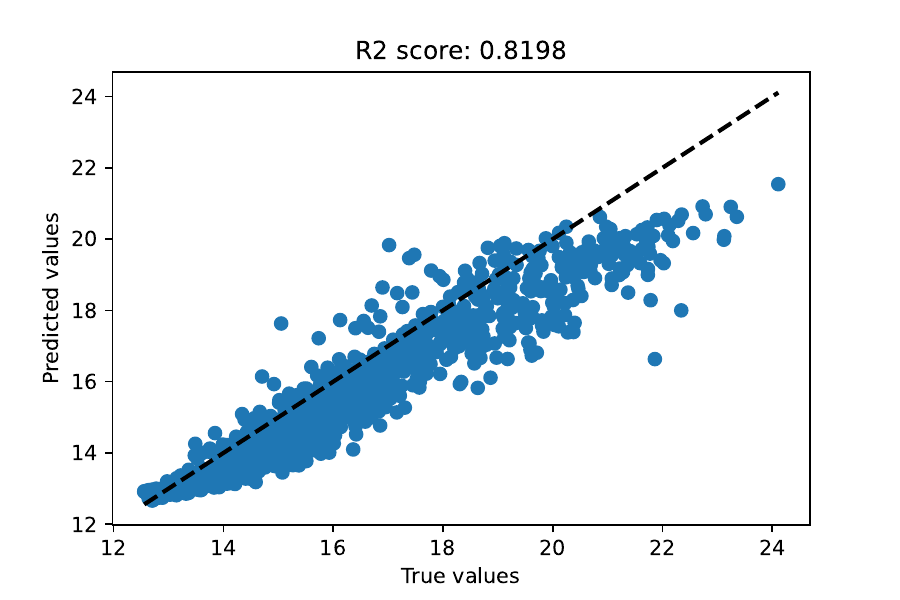}
    \caption{BP}
  \end{subfigure}
  \begin{subfigure}[b]{0.31\textwidth}
    \includegraphics[width=\textwidth]{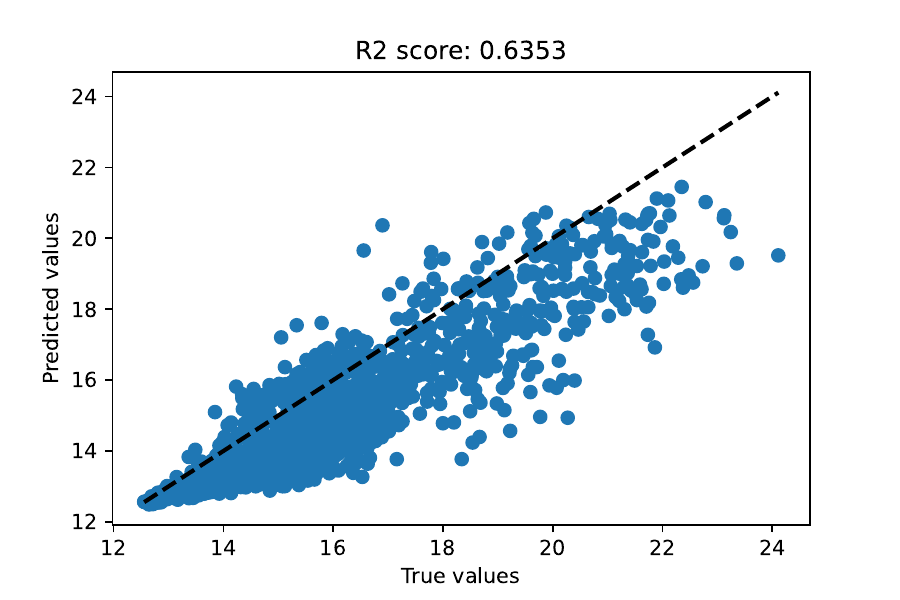}
    \caption{SVR}
  \end{subfigure}
  \begin{subfigure}[b]{0.31\textwidth}
    \includegraphics[width=\textwidth]{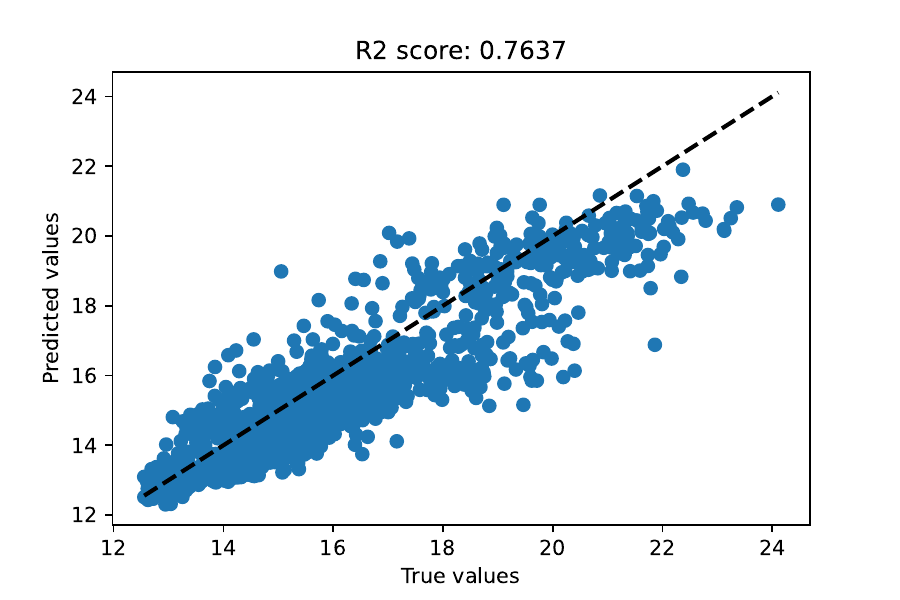}
    \caption{GBDT}
  \end{subfigure}
  \begin{subfigure}[b]{0.31\textwidth}
    \includegraphics[width=\textwidth]{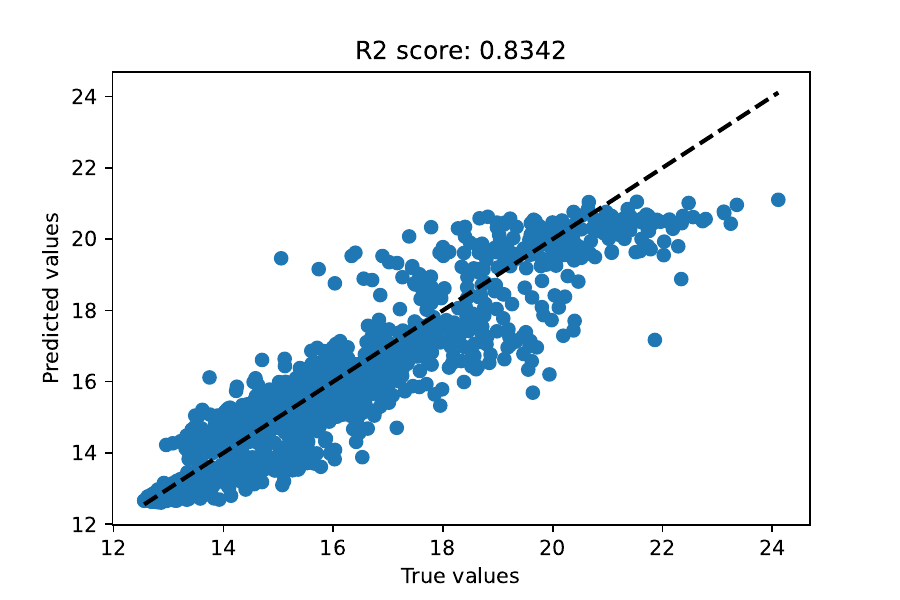}
    \caption{RF}
  \end{subfigure}
  \begin{subfigure}[b]{0.31\textwidth}
    \includegraphics[width=\textwidth]{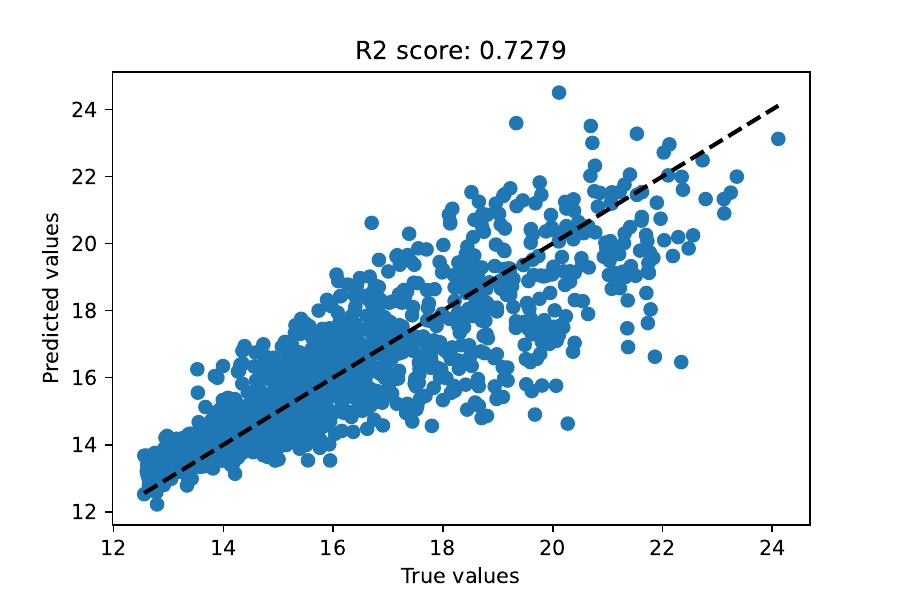}
    \caption{ARIME}
  \end{subfigure}
  \caption{Scatterplot comparing actual and forecasted line loss rates for a 1-hour forecast horizon across different algorithms.}
  \label{fig:scatterplot}
\end{figure*}

\subsection{Experimental Results and Discussion}

In Table~\ref{Tab:results}, we provide a comprehensive analysis of the performance of our proposed Attention-GCN-LSTM model compared to other baseline methods for line loss rate prediction. The evaluation covers various time horizons, including 1 hour, 3 hours, 8 hours, 24 hours, and 168 hours. To ensure a fair comparison, all baseline models (excluding ARIMA) use the same input variables as our proposed model, consisting of 21 variables.

Additionally, Fig.\ref{fig:scatterplot} showcases a detailed scatterplot comparison between the forecasted line loss rates and the actual values specifically for the 1-hour horizon. The scatterplot vividly illustrates the close alignment between the predicted and actual values, with our proposed Attention-GCN-LSTM model consistently outperforming the other algorithms. This visual representation provides valuable insights into the model's ability to capture the patterns and fluctuations of line loss rates accurately.

Among the baseline models, the ARIMA model exhibits the poorest performance due to its limitations in considering the impact of electrical and weather characteristics on line losses, as well as handling complex non-stationary time series data. The RF and GBDT models, as ensemble learning methods, have limited capabilities in expressing non-linear relationships and fail to capture the intricate influencing factors and complex topologies present in power distribution networks. SVR, despite utilizing Support Vector Machine (SVM) for regression, falls short compared to our proposed model, particularly in short-term predictions. BP neural networks have limitations such as the requirement for a large number of samples and weak generalization ability for medium to long-term complex prediction problems.

While LSTM successfully addresses long-term dependencies in RNNs, it still faces challenges in handling longer sequence data and capturing complex relationships in high-dimensional data. Bi-LSTM performs better than LSTM in our experiments, particularly for the 168-hour prediction horizon, but relies heavily on hyperparameter tuning and a larger training data set. TCN-LSTM combines TCN and LSTM to capture both short-term and long-term dependencies, but its performance decreases with longer prediction horizons, indicating potential limitations in capturing the complexities of line loss rate data over longer time frames. Similarly, DBN model also fall short in achieving optimal performance for long-term predictions due to challenges such as capturing complex temporal dependencies and insufficient utilization of sequential information.

PE-CNN, short for Position Encoding Convolutional Neural Network, stands out among the baseline models with its strong performance across all prediction horizons. It incorporates a position encoding scheme that enhances sequential information encoded by a Convolutional Neural Network (CNN). The position encoding captures the temporal dynamics of the line loss rate data, allowing the model to effectively capture patterns and relationships over time. This additional encoding scheme improves the model's ability to capture temporal dependencies and contributes to its strong performance in line loss rate prediction. Despite its notable performance, PE-CNN still falls short of our proposed Attention-GCN-LSTM model. In comparison to our proposed model, PE-CNN exhibits a 28.77\% increase in Root Mean Squared Error (RMSE) and a 30.18\% increase in Mean Absolute Error (MAE), as well as a 12.81\% decrease in the $R2$ score.

\begin{figure*}
\centering
\includegraphics[scale=0.4]{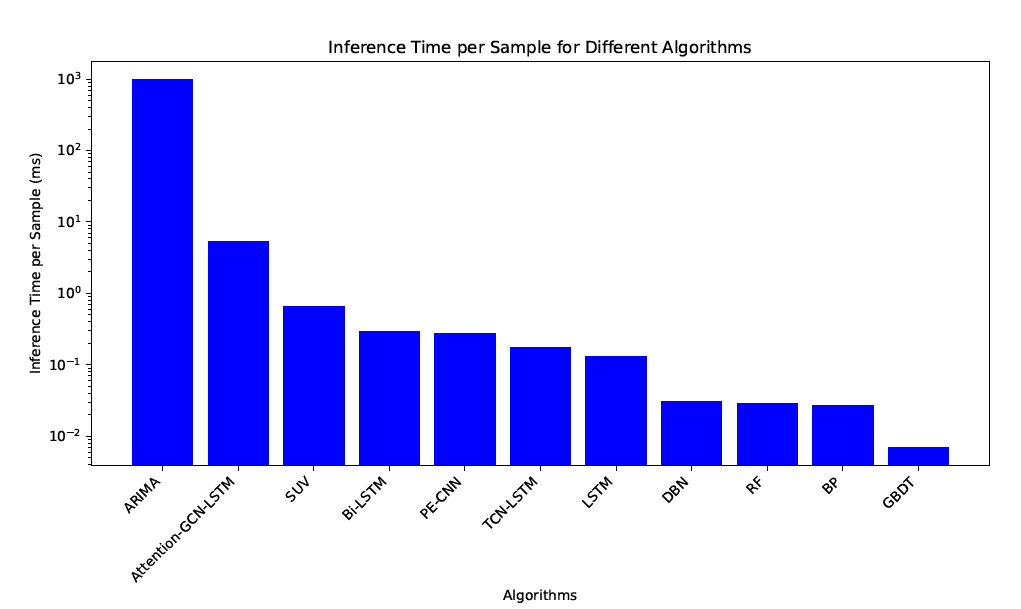}

\caption{Average Inference Time per Sample across Different Algorithms for All Prediction Horizons.}
\label{fig:runtime}
\end{figure*}

As depicted in Fig. \ref{fig:runtime}, The Attention-GCN-LSTM model exhibits a higher inference time of 5.2470 ms compared to other models such as DBN, TCN-LSTM, and PE-CNN. This indicates a greater computational requirement for the model's predictions. However, it is essential to consider the trade-off between inference time and prediction accuracy. Despite the longer inference time, the Attention-GCN-LSTM model achieves superior accuracy compared to other models. This suggests that the incorporation of the GCN component and the joint modeling of various components contribute to the model's enhanced performance, thereby justifying the increased computational cost.

Overall, our proposed Attention-GCN-LSTM model demonstrates significant advantages over the baseline methods, reaffirming its effectiveness in line loss rate prediction. By effectively integrating graph convolution, attention mechanisms, and LSTM, our model achieves superior performance across various evaluation metrics and prediction horizons, demonstrating its potential for practical applications in power distribution network management and planning.

\subsection{Ablation Studies}
To gain deeper insights into the contributions of each component in our proposed Attention-GCN-LSTM model, we conducted ablation studies by progressively excluding different components and observing the resulting degradation in performance.

First, we disabled the Time-level attention (T\_atten) component, which led to a degradation of the $R^2$ score by 5.60\%. This finding highlights the importance of Time-level attention in capturing crucial temporal information and improving the overall prediction performance of the model.

Next, we removed the LSTM module from our model, resulting in a further decrease in performance by 9.5\%. This indicates that the LSTM module plays a critical role in capturing dynamic changes in temporal dependencies, which are essential for accurate line loss rate prediction.

Further, we gradually removed the Distribution transformer-level attention (D\_atten) and Feature-level attention (F\_atten) components from our model. The removal of these attention mechanisms led to continued degradation in performance, indicating their significance in extracting spatial information and enhancing the model's overall performance.

The removal of Distribution transformer-level attention resulted in a degradation in the $R^2$ score by 12.29\%, indicating its role in capturing the relationships and dependencies between different transformers in the distribution network. This attention mechanism enables the model to effectively utilize spatial information and leverage the interactions among transformers for accurate line loss rate prediction.

Similarly, the removal of Feature-level attention led to a degradation in the $R^2$ score by 9.43\%. This demonstrates the importance of Feature-level attention in focusing on important spatial features and capturing their relevance to line loss rates. By attending to specific features, the model can better understand the spatial characteristics and their impact on the prediction task.

The ablation studies summarized in Table ~\ref{Tab:ablation} further highlight the contributions of each component in our proposed Attention-GCN-LSTM model. Time-level attention captures crucial temporal information, LSTM module captures dynamic changes in temporal dependencies, and Distribution transformer-level attention and Feature-level attention extract spatial information and enhance the model's performance by capturing relationships among transformers and relevant spatial features.

\begin{table*}
\centering
 \caption{Ablation experiment results with a prediction horizon of 3 hours. The table presents the performance metrics of our proposed Attention-GCN-LSTM model and its variations obtained through ablation studies. The Architecture column describes the components gradually removed from the original model to observe their impact on performance.}
\begin{tabular}{|c|lll|}
\hline

\multirow{2}{*}{\textbf{Architecture}}            & \multicolumn{3}{c|}{\textbf{Metric}}                                                                           \\ \cline{2-4}
                                                  & \multicolumn{1}{c|}{\textbf{RMSE}}   & \multicolumn{1}{c|}{\textbf{MAE}}    & \multicolumn{1}{c|}{ \begin{math}\mathbf{R^{2}}\end{math}} \\ \hline
\multicolumn{1}{|l|}{\textbf{Our Attention-GCN-LSTM}} & \multicolumn{1}{l|}{\textbf{0.7490}} & \multicolumn{1}{l|}{\textbf{0.5024}} & \textbf{0.8734}                  \\ \hline
\multicolumn{1}{|l|}{\textbf{\qquad- T\_atten (Time-level attention)}}                          & \multicolumn{1}{l|}{0.9701}          & \multicolumn{1}{l|}{0.7148}          & 0.8245                           \\ \hline
\multicolumn{1}{|l|}{\textbf{\qquad\qquad- LSTM (Long Short-Term Memory)}}                               & \multicolumn{1}{l|}{1.0585}          & \multicolumn{1}{l|}{0.8032}          & 0.7910                          \\ \hline
\multicolumn{1}{|l|}{\textbf{\qquad\qquad\qquad- F\_atten (Feature-level attention)}}                               & \multicolumn{1}{l|}{1.0882}          & \multicolumn{1}{l|}{0.8275}          & 0.7791                          \\ \hline
    \multicolumn{1}{|l|}{\textbf{\qquad\qquad\qquad\qquad- D\_atten (Distribution transformer-level attention)}}                               & \multicolumn{1}{l|}{1.1566}          & \multicolumn{1}{l|}{0.9241}          & 0.7505                           \\ \hline
\end{tabular}
\label{Tab:ablation}
\end{table*}

\section{Conclusion}
In this study, we proposed the Attention-GCN-LSTM model, which combines Graph Convolutional Networks (GCN), Long Short-Term Memory (LSTM), and a three-level attention mechanism to accurately forecast line loss rates in power distribution networks. By leveraging the spatio-temporal dependencies of electrical and non-electrical characteristic parameters among distribution transformers, our model aims to enhance the accuracy of line loss rate predictions across multiple time horizons in the short term.

Through comprehensive experimentation and comparison with state-of-the-art baselines, we have demonstrated the superior performance and robustness of our proposed Attention-GCN-LSTM model. It outperforms other models in terms of prediction accuracy and exhibits consistent performance across different prediction horizons. The accurate forecasting of line loss rates in distribution networks is of significant value for power grid planning and operation, as it provides insights into future loss fluctuation ranges.

Moreover, the success of our proposed model suggests its potential application in other spatio-temporal tasks, such as load forecasting. The attention mechanism and integration of GCN and LSTM enable the model to capture complex dependencies and extract meaningful features from high-dimensional data, making it adaptable to a range of predictive tasks.

In summary, this study contributes to the field of line loss prediction in power distribution networks by introducing the Attention-GCN-LSTM model, which achieves superior performance and robustness. Future research can focus on further enhancing the model's ability to handle longer-term predictions and exploring its application in other related domains. Additionally, investigating the interpretability of the model's predictions and understanding the underlying factors contributing to line losses would be valuable directions for future investigations.

\bibliographystyle{unsrtnat}
\bibliography{Axiv_GCN_LSTM}

\end{document}